\pdfoutput=1
\documentclass[11pt]{article}

\usepackage[preprint]{acl}

\usepackage{times}
\usepackage{latexsym}

\usepackage[T1]{fontenc}
\usepackage[utf8]{inputenc}

\usepackage{microtype}

\usepackage{inconsolata}

\usepackage{graphicx}
\usepackage{amsmath}
\usepackage{adjustbox}

\usepackage{hyperref}

\title{\textsc{TopViewRS}: Vision-Language Models as Top-View Spatial Reasoners}

\author{
Chengzu Li\thanks{\ \ Equal contributions. }, 
Caiqi Zhang\footnotemark[1], 
Han Zhou, 
Nigel Collier,
Anna Korhonen,
Ivan Vulić
\\
Language Technology Lab, University of Cambridge \\
\texttt{\{cl917, cz391, hz416, nhc30, alk23, iv250\}@cam.ac.uk} \\
\url{https://topviewrs.github.io}
}

\usepackage{xspace}

\newcommand{\rparagraph}[1]{\vspace{1.2mm}\noindent\textbf{#1.}}

\newcommand{\rrparagraph}[1]{\vspace{0.5mm}\noindent\textit{#1:}}
\newcommand{\sparagraph}[1]{\vspace{0.0mm}\noindent\textbf{#1.}}

\newcommand{\iparagraphnodot}[1]{\vspace{0.0mm}\noindent\textit{#1}}

\definecolor{Gray}{gray}{0.92}
\definecolor{racing-green}{rgb}{0.0, 0.8, 0.6}
\definecolor{awesome-red}{rgb}{1.0, 0.13, 0.32}

\newcommand{\topviewers}{\textsc{TopViewRS}\xspace}

\usepackage{booktabs}
\usepackage{url}
\usepackage{makecell}
\usepackage{enumitem}
\usepackage{caption}
\usepackage{subcaption}
\usepackage{multirow}
\usepackage{colortbl}
\usepackage{svg}
\usepackage{amsmath}
\usepackage{fancyvrb}
\usepackage{spverbatim}
\usepackage{subcaption}
\usepackage{tcolorbox}

\tcbuselibrary{skins, breakable, theorems}

\begin{document}
\maketitle

\begin{figure*}[t!]
    \centering
    \includegraphics[width=0.985\textwidth]{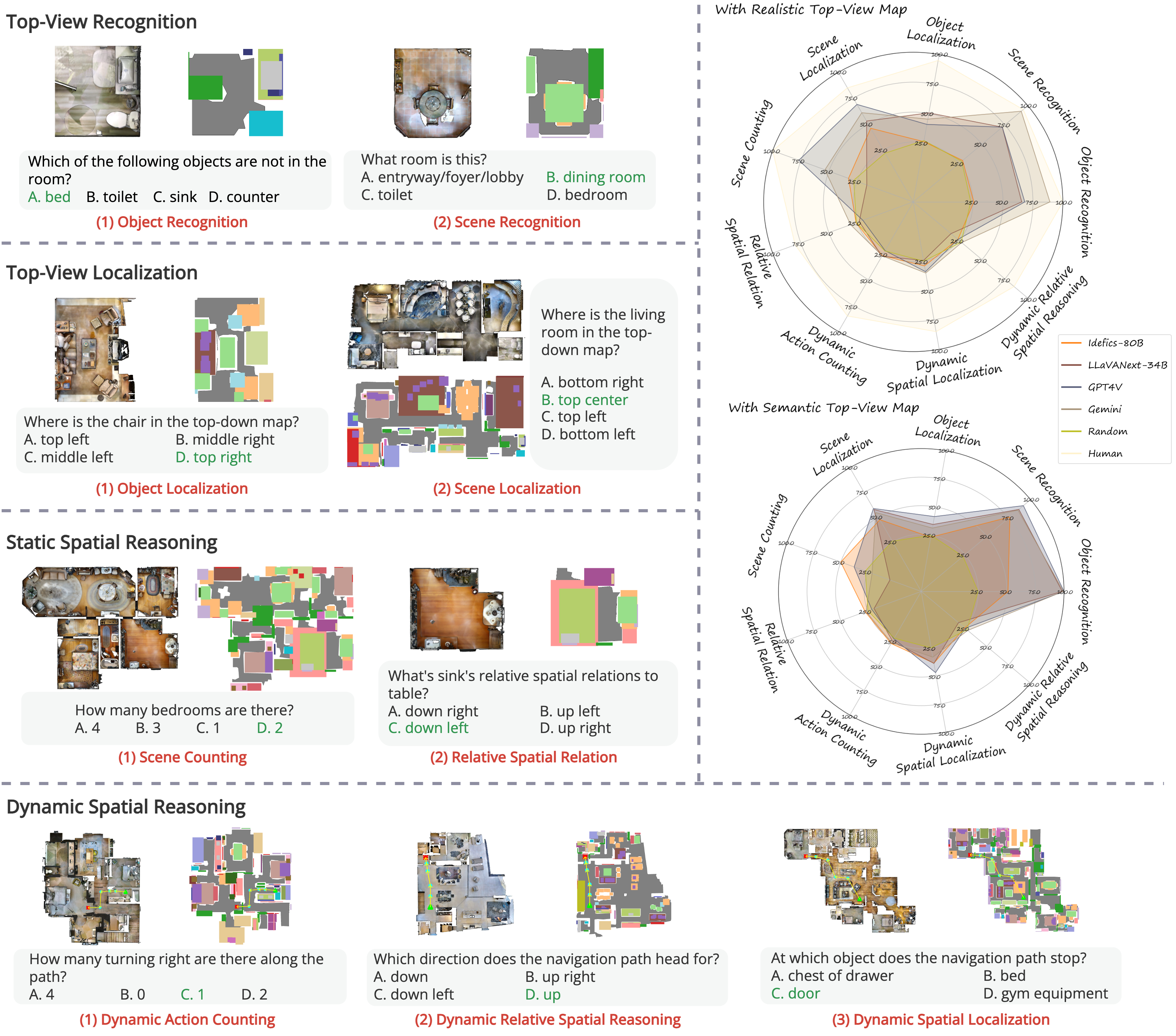}
    \caption{Illustration of the four evaluation tasks with an incremental level of complexity on the two types of top-view maps (photo-realistic versus semantic maps), covering top-view perception and spatial reasoning abilities, with 9 sub-tasks in total (red font), focusing on different, well-defined VLM abilities. The radar graphs (top right) compare the representative models' performance on all sub-tasks, indicating \textit{a large gap with human performance}.}
    \label{fig:front fig}
    \vspace{-6mm}
\end{figure*}

\begin{abstract}

Top-view perspective denotes a typical way in which humans read and reason over different types of maps, and it is vital for localization and navigation of humans as well as of `non-human' agents, such as the ones backed by large Vision-Language Models (VLMs). Nonetheless, spatial reasoning capabilities of modern VLMs remain unattested and underexplored. In this work, we thus study their capability to understand and reason over spatial relations from the top view. The focus on top view also enables controlled evaluations at different granularity of spatial reasoning; we clearly disentangle different abilities (e.g., recognizing particular objects versus understanding their relative positions). We introduce the \topviewers (\textbf{Top}-\textbf{View} \textbf{R}easoning in \textbf{S}pace) dataset, consisting of 11,384 multiple-choice questions with either realistic or semantic top-view map as visual input. We then use it to study and evaluate VLMs across 4 perception and reasoning tasks with different levels of complexity. Evaluation of 10 representative open- and closed-source VLMs reveals the \textit{gap} of more than 50\% compared to average human performance, and it is even \textit{lower} than the random baseline in some cases. Although additional experiments show that Chain-of-Thought reasoning can boost model capabilities by 5.82\% on average, the overall performance of VLMs remains limited. Our findings underscore the critical need for enhanced model capability in top-view spatial reasoning and set a foundation for further research towards human-level proficiency of VLMs in real-world multimodal tasks.
\end{abstract}

\section{Introduction}

Large Language Models (LLMs) such as Llama 2 and 3~\citep{touvron2023llama}, Mistral~\citep{jiang2023mistral}, and GPT models \citep{gpt3.5}  have delivered impressive performance across a range of text-based tasks and applications such as question answering, language generation, and arithmetic reasoning \citep{qin-etal-2023-chatgpt, zhao2023survey}.  Building on these text-only LLMs, the so-called Vision Language Models (VLMs), equipped with the capability to process and reason over multimodal vision-language information, have enabled multi-modal processing~\citep{yin2023survey, wu2023multimodal}. They ground language reasoning ability of LLMs into the information of different modalities~\cite{chandu-etal-2021-grounding}. 
Prominent examples of VLMs such as LLaVA~\citep{liu2023llava}, GPT-4V~\citep{openai2023gpt4v}, and Gemini~\citep{geminiteam2024gemini}, have demonstrated strong performance across applications such as visual question answering \cite{li2023comprehensive}, image captioning \cite{diesendruck2024learning}, and object grounding \cite{zheng2024gpt4vision}. 

Spatial reasoning, as one of the fundamental desirable properties of and requirements for VLMs, has also gained increased attention recently \citep{rajabi2023grounded, liu-etal-2023-visual, chen2024spatialvlm}. It requires grounding the model's reasoning ability with natural language into its visual perception of the surrounding environment \cite{Freksa1991}. In particular, it involves two critical steps: (i) \textit{interpreting} the environment visually, and (ii) \textit{reasoning} over spatial relations. As a fundamental ability for the model to recognize, understand, and navigate through the physical world, it plays a crucial role in various downstream tasks such as vision-language generation~\cite{li2024semantic} and embodied AI~\cite{cho2024spatiallyaware}.

However, previous research has focused on exploring spatial reasoning abilities of VLMs only from a conventional first-person perspective view \citep{liu-etal-2023-visual}. In this work, we aim to study and evaluate spatial understanding and reasoning capability of VLMs from the \textit{top-view perspective}, also referred to as the bird's-eye view \citep{bev_li_2024}. When compared to the conventional perspective view, top view offers better \textit{natural alignment}: it is a typical view to read different maps or present, e.g., floor plans. Moreover, it is inherently more \textit{complex}: top-view maps encapsulate a wealth of information about different scenes, locations, objects and their relationships in the environment based on a single image. In addition to the \textit{photo-realistic} top-view maps, \textit{semantic} top-view maps \cite{Nanwani_2023, li2024semantic} use different colors to represent different types of objects; we run experiments with both map types, see Figure~\ref{fig:front fig}.

One advantage of top-view maps is that they define a controlled and interpretable experimental framework. Indoor scenes, which are the focus of this work, typically feature a relatively stable set of objects and layouts, making them ideal for controlled studies. This allows us to disentangle and investigate various aspects of spatial reasoning and VLMs' capabilities in a controlled manner.\footnote{For instance, we can apply different interventions (e.g., drawing a navigation trajectory in a realistic map, or changing the color-object mapping in a semantic top-view map).}

In this work, we thus investigate the basic top-view spatial understanding and reasoning abilities of current state-of-the-art VLMs across four tasks of gradually increasing complexity, and their finer-grained sub-tasks. The tasks are as follows.
\textit{1) Top-View Recognition} assesses whether the model can recognize concrete objects and scenes in top-view maps. 
\textit{2) Top-View Localization} evaluates the ability to localize objects or regions on a map based on textual descriptions. 
\textit{(3) Static Spatial Reasoning} investigates whether the model can reason about spatial relationships among localized objects and regions within the map. 
\textit{(4) Dynamic Spatial Reasoning} evaluates reasoning about spatial relations along the points of a dynamic navigation path.
Figure \ref{fig:front fig} illustrates all the tasks with concrete examples. As one key finding of this study, conducted evaluations reveal that current VLMs lack sufficient capability to effectively tackle top-view spatial reasoning challenges, indicating substantial room for improvement in future research.

\rparagraph{Contributions}
    \textbf{1)} We define the top-view spatial reasoning challenge for VLMs via 4 carefully designed tasks of increasing complexity, also encompassing 9 distinct fine-grained sub-tasks with a structured design of the questions focusing on different model abilities. 
    \textbf {2)} We collect the \topviewers dataset, comprising 11,384 multiple-choice questions with either photo-realistic or semantic top-view maps of real-world scenarios through a pipeline of automatic collection followed by human alignment. 
    \textbf{3)} We use \topviewers to evaluate and study 10 VLMs from different model families and sizes, highlighting the performance gap compared to human annotators \footnote{We release the code in \url{https://github.com/cambridgeltl/topviewrs}. }. 
\section{Related Work}
\label{sec:related work}

\rparagraph{Top-View Map Understanding}
There are only limited studies in NLP focused on the use of top-view maps, though considerable research has been conducted within the broader AI community on the so-called \textit{bird's-eye view}, which is an instance of top view. 
This body of work has explored applications in autonomous driving \cite{unger2023multicamera, Li_2024_bev}, with contributions on fusing different types of views~\cite{Qin_2023_ICCV} and working with arbitrary camera setups~\cite{Peng_2023}. 
In other application scenarios, \citet{Yan_2021_ICCV} introduce a bird's-eye view person re-identification task with 114k images of the person. 

Efforts to bridge top-view images with natural language in applications beyond the above are less diverse.
The WAY dataset, proposed by \citet{hahn-etal-2020-localization}, contains 6,154 dialogs aimed at localizing an observer’s position on a top-view map through conversations between an observer and a locator.
This dataset has inspired follow-up research focusing on merging vision with dialog information \cite{zhang2024dialoc} and leveraging pretraining strategies to enhance performance~\cite{hahn-rehg-2022-transformer}. In general, prior research does not assess VLMs' basic spatial reasoning abilities with top-view images and lacks fine-grained and controllable analysis of these fundamental abilities. 

\rparagraph{Spatial Reasoning on Multi-Modal Vision-Text}
There has been a body of work on text-only spatial reasoning with the advancement of LLMs \cite{yamada2024evaluating}, within the context of relative spatial relation recognition \cite{mirzaee-etal-2021-spartqa, stepGame2022shi}, natural language navigation \cite{yamada2024evaluating}, and planning \cite{NEURIPS2023_dc9d5dcf} (see Appendix \ref{app:related_work} for a more complete overview). 

Cross-modal spatial reasoning puts forward higher requirements for the models in terms of language grounding~\cite{rozanova2021grounding, rajabi2023grounded}. \citet{liu-etal-2023-visual} investigate spatial reasoning with 2D natural realistic front view images and \citet{chen2024spatialvlm} extend the analysis to 3D point clouds. 
The environmental contexts become more diverse compared to synthetic symbols in text-only spatial reasoning, ranging from indoor environments \cite{koch2024open3dsg} to outdoor street views \cite{Chen_2019}. 
Regarding typical tasks, visual QA (VQA) is the mainstream task for benchmarking spatial reasoning abilities \cite{dong-etal-2021-visually, Banerjee_2021_ICCV, liu-etal-2023-visual, li2023seedbench, Li_2023, kamath-etal-2023-whats}, while other tasks include vision-Language navigation \cite{Chen_2019, li2024semantic} and user interface grounding \cite{rozanova2021grounding}.\footnote{Research on multi-modal spatial reasoning also intersects with efforts from the computer vision community on scene understanding \cite{8099827}, simultaneous localization and mapping~\cite{Cadena_2016}, and combining LLMs with representations of the 3D physical world~\cite{hong20233dllm}.}

We stress that none of the prior research efforts allows for \textit{disentangled evaluation} of models' spatial reasoning abilities. Prior work typically conflates object recognition with spatial reasoning. We thus design a dataset and conduct a study that not only offers insight into fundamental abilities but also allows for easier interpretation of results (\S\ref{sec:dataset}).

\section{Task Definition}
\label{sec:task}
Following prior work~\cite{li2023seedbench}, we frame all tasks as multiple-choice QA tasks. Given a top-view (realistic or semantic) map of a room $M$, the model must choose the correct option $o_i$ from the four options provided $O = \{o_0, o_1, o_2, o_3\}$ that answers the question.\footnote{For simplicity, for each question, there is always a \textit{single correct answer}.} This format simplifies the evaluation and interpretation of the results.

\rparagraph{Top-View Maps} We provide two different types of top-view maps to the models: realistic maps $M_{\text{Real}}$ and semantic maps $M_{\text{Sem}}$. 
Realistic maps are constructed by placing a simulated orthographic camera above the scene to capture a photo-realistic top-view image. 
Semantic maps represent objects in the scene with colored bounding boxes. Each object is assigned a specific color and labeled at the same relative coordinates on the map to preserve the object's semantic information and spatial allocation. 
In comparison to realistic maps, semantic maps simplify the initial step of spatial reasoning (i.e., environment interpretation) by labeling the object types with corresponding colors and excluding irrelevant additional details such as shape and texture found in realistic top-view maps.
Given the customizable and flexible nature of color-object mapping, the semantic map can also serve as an ideal testbed for evaluating models' out-of-distribution (OOD) performance, thereby encouraging further exploration beyond the scope of this work. 
Example maps are in Figure \ref{fig:front fig}.

\rparagraph{Tasks and Sub-Tasks} We define 4 different tasks which cover a total of 9 finer-grained sub-tasks, with concrete examples shown in Figure \ref{fig:front fig}. The tasks are designed to have an increasing level of complexity, where each subsequent task depends on the abilities measured in the preceding one(s). 

\iparagraphnodot{\textbf{(1) Top-View Recognition}} evaluates the fundamental ability to interpret the input map, and covers two sub-tasks: \textit{Object Recognition} and \textit{Scene Recognition}. 
It does not require the model to identify specific locations of objects and rooms.

\iparagraphnodot{\textbf{(2) Top-View Localization}} investigates whether the model can localize objects or rooms in the top-view map based on textual descriptions, including \textit{Object Localization} and \textit{Scene Localization} as two sub-tasks. 
Beyond understanding the top-view map as a whole, it requires the model to ground entities in the map, representing the model's ability to align spatial descriptions with corresponding locations.

\iparagraphnodot{\textbf{(3) Static Spatial Reasoning}} aims to evaluate the model's spatial reasoning ability with more complex questions. It includes two sub-tasks: reasoning over \textit{Scene Counting} and \textit{Relative Spatial Relations} between different objects and rooms. These questions require the model to perform multi-step reasoning based on the recognition and localization of entities in the top-view map.

\iparagraphnodot{\textbf{(4) Dynamic Spatial Reasoning.}}
Finally, we introduce a novel task that involves dynamic spatial reasoning over top-view maps in the context of agent navigation. It requires the model to understand the sequential relations along the points of the navigation path (sub-task \textit{Dynamic Action Counting}) and answer spatial questions with regard to the dynamic navigation path (sub-task \textit{Dynamic Relative Spatial Reasoning}) and the circumstantial environments (\textit{Dynamic Spatial Localization}). 
\section{\topviewers Dataset}
\label{sec:dataset}
In order to study and evaluate the abilities of state-of-the-art VLMs on the four tasks spanning 9 sub-tasks from \S\ref{sec:task}, we now introduce a novel evaluation dataset, \topviewers, which focuses on \textit{top-view maps of indoor scenes} (i.e., houses and rooms), discussed in what follows.

\rparagraph{Dataset Features}
It introduces several advancements and innovative features that distinguish it from all prior visual spatial reasoning datasets.

\rrparagraph{1) Multi-Scale Top-View Maps}
The selected top-view maps of indoor scenes (see Figure~\ref{fig:front fig}) provide a more natural representation of spatial environments that aligns with human cognitive map \cite{cognitive_topview}.
This makes benchmarking spatial awareness more straightforward and meanwhile mitigates spurious correlations in the positions between objects commonly found in realistic front-view images. 
Compared to the front view, the multi-scale top-view maps of single rooms and full houses add more divergence in the granularity of the entities (objects or rooms) in spatial reasoning. 
Meanwhile, we provide both realistic maps and semantic maps for more comprehensive evaluation.

\rrparagraph{2) Realistic Environmental Scenarios with Rich Object Sets} 
We provide real-world environments from indoor scenes, with 80 objects per scene on average, ensuring a natural distribution and complexity of object locations. This also sets it apart from existing front-view spatial reasoning datasets, which often contain only a handful of objects. 

\rrparagraph{3) Structured Question Framework} 
Unlike previous datasets \cite{li2023seedbench, liu-etal-2023-visual, kamath-etal-2023-whats}, which often conflate spatial reasoning with object recognition, our dataset clearly defines four tasks including 9 sub-tasks in total using diverse question templates. This structured approach allows for a fine-grained evaluation and analysis of models’ capabilities from various perspectives and levels of granularity.

\begin{figure*}[t!]
    \centering
    \includegraphics[width=0.97\textwidth]{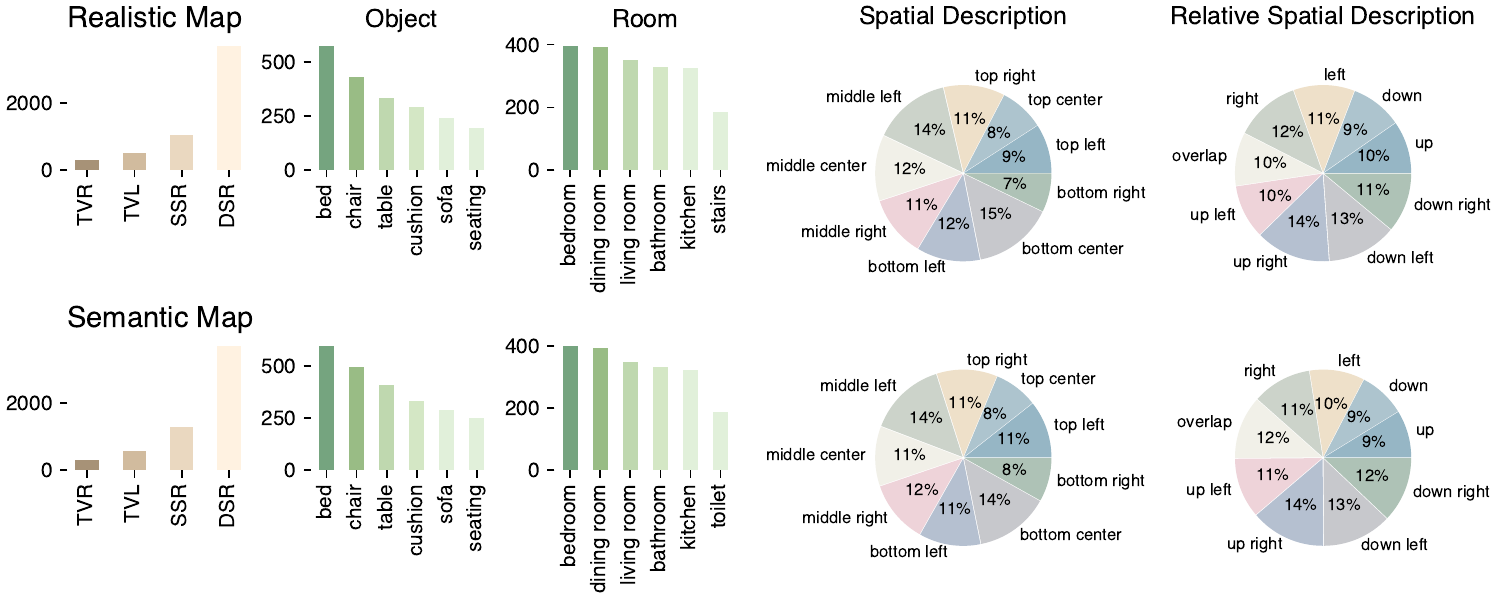}
    \caption{\topviewers data statistics, showing distribution of task sizes, objects, regions, spatial and relative spatial descriptions in realistic and semantic map settings, where the tasks are described with their initials for visualization. }
    \label{fig:dataset statistics}
    \vspace{-4mm}
\end{figure*}

\rparagraph{Dataset Collection}
We employ a two-stage data collection strategy that includes \textit{automatic collection from a simulator} and \textit{alignment through human judgment}. First, to approximate real-life scenarios, we use the Matterport3D dataset~\cite{Matterport3D}, which includes 90 building-scale scenes with instance-level semantic and room-level region annotations in 3D meshes. We filter these to exclude multi-floor and low-quality scenes, selecting 7 scenes with an average of 80 objects and 12 rooms each. Realistic top-view maps are extracted using orthographic cameras, and semantic top-view maps are constructed using the Habitat \cite{habitat19iccv, szot2021habitat} simulation environment. We then design a structured question framework with 15 templates to minimize human labor and standardize the data collection process. 
To ensure quality, a second stage of manual \textit{human judgment} aligns and verifies the data, ensuring questions are natural and correct. Participants are encouraged to discard or modify data points to improve quality, maintaining alignment with human judgments. We refer readers to Appendix \ref{app:dataset construction} for further details regarding the data collection process.

\rparagraph{Dataset Statistics} 
\label{subsec:dataset analysis}
The \topviewers evaluation dataset comprises a total of 11,384 multiple-choice questions after human verification, with 5,539 questions associated with realistic top-view maps, and 5,845 with semantic top-view maps.  Human verification keeps 587/784 questions from the automatic collection phase for Top-View Recognition, 1,077/1,384 for Top-View Localization, 2,340/3,080 for Static Spatial Reasoning. 
The choices are uniformly distributed over choices A (\textit{25.5\%}), B (\textit{24.6\%}), C (\textit{24.5\%}) and D (\textit{25.4\%}). 
Figure~\ref{fig:dataset statistics} shows the distribution of different tasks, objects, regions and spatial descriptions. 
The size of each task aligns with its corresponding difficulty level, where the easier task comprises fewer examples. 
We provide further insights and technical details in Appendix~\ref{appsec:dataset analysis}. 
\section{Experiments and Results}

\sparagraph{Models and Implementation}
We test a representative selection of both open-sourced and close-sourced models which achieve state-of-the-art performance on a range of multimodal benchmarks \cite{liu2023mmbench, li2023seedbench} in a zero-shot inference setup. 
Regarding open-sourced models, we study and evaluate Idefics (9B \& 80B) \cite{laurenon2023obelics}, LLaVA-Next (7B, 13B \& 34B) \cite{liu2024llavanext}, InternLM-XComposer2 (7B) \cite{dong2024internlmxcomposer2}, Qwen-VL (7B) \cite{bai2023qwenvl}.
The chosen close-sourced models are GPT-4V \cite{openai2023gpt4v} and Gemini \cite{geminiteam2024gemini}.\footnote{We use \textit{GPT-4-turbo-2024-04-09} of GPT-4V and latest stable \textit{gemini-pro-vision 1.0} of Gemini.} All the models are implemented within the VLMEvalKit framework~\cite{2023opencompass}.

\rparagraph{Prompts}
For realistic maps, we provide the VLMs with the task description along with the multiple-choice question. For semantic maps, in addition to the information above, we also introduce the concept of a semantic map to the model and provide the color-object mapping in the prompt in order to facilitate its understanding of the abstract map. 
We only provide the color-object mappings of the colors that are presented in the semantic map as a pre-processing strategy in order to exclude irrelevant information. 
For the specific prompting templates used in this paper, we refer to Appendix~\ref{app:prompt}. 

\rparagraph{Evaluation Measures}
We measure multiple-choice QA accuracy via \textit{Exact Match (EM)} and \textit{Partial Match (PM)}. 
EM measures whether the predicted option indices are exactly the same as the label indices. However, there may be cases where the correct answer to the question can be considered partially correct, e.g., the answer is \textit{top right} while the prediction is \textit{top left}.
PM then calculates the proportion of overlapping words between the predicted answer and the gold answer. It is calculated based on the correctness of the text spans (or words) of predicted options, as given by: 

\vspace{-1mm}
{\footnotesize
\begin{equation*}
    PM = \frac{\left| \{\text{labels}\} \cap \{\text{predictions}\} \right|}
    {\max\left(\left| \{\text{labels}\} \right|, \left| \{\text{predictions}\} \right|\right)}
\end{equation*}
}

\subsection{Results and Discussion}
\label{subsec:results}

\begin{table*}[t]
\centering
\renewcommand\arraystretch{1.15}
\resizebox{\textwidth}{!}{%
\begin{tabular}{lccccccccccc}
\hline
\textbf{Model}                              & \multicolumn{1}{l|}{}                           & \multicolumn{2}{c|}{\textbf{Idefics}}                                              & \multicolumn{4}{c|}{\textbf{LLaVANext}}                                                                                                                     & \multicolumn{1}{c|}{\textbf{XComposer2}}           & \multicolumn{1}{c|}{\textbf{Qwen-VL}}              & \multicolumn{1}{c|}{\textbf{GPT-4V}}                         & \textbf{Gemini}                          \\
Model Size                                  & \multicolumn{1}{l|}{}                           & 9B                            & \multicolumn{1}{c|}{80B}                           & vicuna 7B                     & mistral 7B                    & vicuna 13B                    & \multicolumn{1}{c|}{34B}                                    & \multicolumn{1}{c|}{7B}                            & \multicolumn{1}{c|}{7B}                            & \multicolumn{1}{c|}{API}                                    & API                                          \\ \hline
\multicolumn{12}{l}{\cellcolor[HTML]{D9D9D9}\textbf{Realistic Map}}                                                                                                                                                                                                                                                                                                                                                                                                                                                                                                           \\
                                            & \multicolumn{1}{c|}{EM}                         & 41.10                         & \multicolumn{1}{c|}{26.71}                         & 67.47                         & 61.30                         & 61.64                         & \multicolumn{1}{c|}{67.81}                                  & \multicolumn{1}{c|}{37.67}                         & \multicolumn{1}{c|}{27.05}                         & \multicolumn{1}{c|}{69.52}                                  & \textbf{90.41}                               \\
\multirow{-2}{*}{Top-View Recognition}    & \multicolumn{1}{c|}{\cellcolor[HTML]{EFEFEF}PM} & \cellcolor[HTML]{EFEFEF}41.10 & \multicolumn{1}{c|}{\cellcolor[HTML]{EFEFEF}26.88} & \cellcolor[HTML]{EFEFEF}67.64 & \cellcolor[HTML]{EFEFEF}61.47 & \cellcolor[HTML]{EFEFEF}61.99 & \multicolumn{1}{c|}{\cellcolor[HTML]{EFEFEF}67.81}          & \multicolumn{1}{c|}{\cellcolor[HTML]{EFEFEF}37.67} & \multicolumn{1}{c|}{\cellcolor[HTML]{EFEFEF}27.26} & \multicolumn{1}{c|}{\cellcolor[HTML]{EFEFEF}69.86}          & \cellcolor[HTML]{EFEFEF}\textbf{90.58}       \\
                                            & \multicolumn{1}{c|}{EM}                         & 30.39                         & \multicolumn{1}{c|}{30.00}                         & 42.16                         & 33.92                         & 41.18                         & \multicolumn{1}{c|}{\textbf{50.98}}                         & \multicolumn{1}{c|}{27.84}                         & \multicolumn{1}{c|}{16.27}                         & \multicolumn{1}{c|}{46.27}                                  & 48.24                                        \\
\multirow{-2}{*}{Top-View Localization}     & \multicolumn{1}{c|}{\cellcolor[HTML]{EFEFEF}PM} & \cellcolor[HTML]{EFEFEF}46.42 & \multicolumn{1}{c|}{\cellcolor[HTML]{EFEFEF}46.08} & \cellcolor[HTML]{EFEFEF}56.67 & \cellcolor[HTML]{EFEFEF}48.63 & \cellcolor[HTML]{EFEFEF}54.31 & \multicolumn{1}{c|}{\cellcolor[HTML]{EFEFEF}\textbf{61.76}} & \multicolumn{1}{c|}{\cellcolor[HTML]{EFEFEF}41.86} & \multicolumn{1}{c|}{\cellcolor[HTML]{EFEFEF}26.31} & \multicolumn{1}{c|}{\cellcolor[HTML]{EFEFEF}60.39}          & \cellcolor[HTML]{EFEFEF}60.98                \\
                                            & \multicolumn{1}{c|}{EM}                         & 24.07                         & \multicolumn{1}{c|}{26.07}                         & 19.87                         & 24.36                         & 20.25                         & \multicolumn{1}{c|}{22.73}                                  & \multicolumn{1}{c|}{25.79}                         & \multicolumn{1}{c|}{14.71}                         & \multicolumn{1}{c|}{22.16}                                  & \multicolumn{1}{c}{\textbf{31.61}}                         \\
\multirow{-2}{*}{Static Spatial Reasoning}         & \multicolumn{1}{c|}{\cellcolor[HTML]{EFEFEF}PM} & \cellcolor[HTML]{EFEFEF}33.68 & \multicolumn{1}{c|}{\cellcolor[HTML]{EFEFEF}38.52} & \cellcolor[HTML]{EFEFEF}34.40  & \cellcolor[HTML]{EFEFEF}37.34 & \cellcolor[HTML]{EFEFEF}36.26 & \multicolumn{1}{c|}{\cellcolor[HTML]{EFEFEF}35.56}          & \multicolumn{1}{c|}{\cellcolor[HTML]{EFEFEF}38.73} & \multicolumn{1}{c|}{\cellcolor[HTML]{EFEFEF}21.15} & \multicolumn{1}{c|}{\cellcolor[HTML]{EFEFEF}35.59}          & \multicolumn{1}{c}{\cellcolor[HTML]{EFEFEF}\textbf{45.22}} \\
                                            & \multicolumn{1}{c|}{EM}                         & 38.10                         & \multicolumn{1}{c|}{27.94}                         & \textbf{38.81}                         & 24.31                         & 29.08                         & \multicolumn{1}{c|}{23.79}                                  & \multicolumn{1}{c|}{24.07}                         & \multicolumn{1}{c|}{22.11}                         & \multicolumn{1}{c|}{30.29}                                  & \multicolumn{1}{c}{32.60}                         \\
\multirow{-2}{*}{Dynamic Spatial Reasoning} & \multicolumn{1}{c|}{\cellcolor[HTML]{EFEFEF}PM} & \cellcolor[HTML]{EFEFEF}40.88 & \multicolumn{1}{c|}{\cellcolor[HTML]{EFEFEF}30.68} & \cellcolor[HTML]{EFEFEF}\textbf{42.15} & \cellcolor[HTML]{EFEFEF}26.69 & \cellcolor[HTML]{EFEFEF}32.89 & \multicolumn{1}{c|}{\cellcolor[HTML]{EFEFEF}27.28}          & \multicolumn{1}{c|}{\cellcolor[HTML]{EFEFEF}26.86} & \multicolumn{1}{c|}{\cellcolor[HTML]{EFEFEF}24.65} & \multicolumn{1}{c|}{\cellcolor[HTML]{EFEFEF}33.86}          & \multicolumn{1}{c}{\cellcolor[HTML]{EFEFEF}35.80} \\ \hline
\multicolumn{12}{l}{\cellcolor[HTML]{D9D9D9}\textbf{Semantic Map}}                                                                                                                                                                                                                                                                                                                                                                                                                                                                                                      \\
                                            & \multicolumn{1}{c|}{EM}                         & 60.68                         & \multicolumn{1}{c|}{59.32}                         & 88.81                         & 80.00                            & 88.14                         & \multicolumn{1}{c|}{94.58}                                  & \multicolumn{1}{c|}{43.05}                         & \multicolumn{1}{c|}{19.66}                         & \multicolumn{1}{c|}{\textbf{97.29}}                         & 94.92                                        \\
\multirow{-2}{*}{Top-View Recognition}    & \multicolumn{1}{c|}{\cellcolor[HTML]{EFEFEF}PM} & \cellcolor[HTML]{EFEFEF}60.68 & \multicolumn{1}{c|}{\cellcolor[HTML]{EFEFEF}59.32} & \cellcolor[HTML]{EFEFEF}88.81 & \cellcolor[HTML]{EFEFEF}80.00    & \cellcolor[HTML]{EFEFEF}88.49 & \multicolumn{1}{c|}{\cellcolor[HTML]{EFEFEF}94.58}          & \multicolumn{1}{c|}{\cellcolor[HTML]{EFEFEF}43.05} & \multicolumn{1}{c|}{\cellcolor[HTML]{EFEFEF}20.05} & \multicolumn{1}{c|}{\cellcolor[HTML]{EFEFEF}\textbf{97.29}} & \cellcolor[HTML]{EFEFEF}94.92                \\
                                            & \multicolumn{1}{c|}{EM}                         & 31.21                         & \multicolumn{1}{c|}{27.34}                         & 25.40                         & 32.10                         & 17.28                         & \multicolumn{1}{c|}{38.45}                                  & \multicolumn{1}{c|}{24.87}                         & \multicolumn{1}{c|}{9.70}                          & \multicolumn{1}{c|}{\textbf{44.44}}                         & 35.27                                        \\
\multirow{-2}{*}{Top-View Localization}     & \multicolumn{1}{c|}{\cellcolor[HTML]{EFEFEF}PM} & \cellcolor[HTML]{EFEFEF}47.62 & \multicolumn{1}{c|}{\cellcolor[HTML]{EFEFEF}45.41} & \cellcolor[HTML]{EFEFEF}44.27 & \cellcolor[HTML]{EFEFEF}47.80 & \cellcolor[HTML]{EFEFEF}23.66 & \multicolumn{1}{c|}{\cellcolor[HTML]{EFEFEF}53.79}          & \multicolumn{1}{c|}{\cellcolor[HTML]{EFEFEF}41.09} & \multicolumn{1}{c|}{\cellcolor[HTML]{EFEFEF}13.99} & \multicolumn{1}{c|}{\cellcolor[HTML]{EFEFEF}\textbf{58.55}} & \cellcolor[HTML]{EFEFEF}49.91                \\
                                            & \multicolumn{1}{c|}{EM}                         & 23.82                         & \multicolumn{1}{c|}{\textbf{28.07}}                & 18.72                         & 24.28                         & 16.63                         & \multicolumn{1}{c|}{18.41}                                  & \multicolumn{1}{c|}{23.05}                         & \multicolumn{1}{c|}{14.85}                         & \multicolumn{1}{c|}{21.73}                                  & 26.22                                        \\
\multirow{-2}{*}{Static Spatial Reasoning}         & \multicolumn{1}{c|}{\cellcolor[HTML]{EFEFEF}PM} & \cellcolor[HTML]{EFEFEF}34.13 & \multicolumn{1}{c|}{\cellcolor[HTML]{EFEFEF}38.17} & \cellcolor[HTML]{EFEFEF}30.57 & \cellcolor[HTML]{EFEFEF}37.26 & \cellcolor[HTML]{EFEFEF}29.94 & \multicolumn{1}{c|}{\cellcolor[HTML]{EFEFEF}31.22}          & \multicolumn{1}{c|}{\cellcolor[HTML]{EFEFEF}35.50}  & \multicolumn{1}{c|}{\cellcolor[HTML]{EFEFEF}21.99} & \multicolumn{1}{c|}{\cellcolor[HTML]{EFEFEF}33.09}          & \cellcolor[HTML]{EFEFEF}\textbf{39.12}       \\
                                            & \multicolumn{1}{c|}{EM}                         & 36.67                         & \multicolumn{1}{c|}{34.55}                         & 37.45                         & 26.23                         & 19.92                         & \multicolumn{1}{c|}{33.12}                                  & \multicolumn{1}{c|}{21.60}                          & \multicolumn{1}{c|}{23.55}                         & \multicolumn{1}{c|}{\textbf{39.30}}                          & 31.41                                        \\
\multirow{-2}{*}{Dynamic Spatial Reasoning} & \multicolumn{1}{c|}{\cellcolor[HTML]{EFEFEF}PM} & \cellcolor[HTML]{EFEFEF}39.92 & \multicolumn{1}{c|}{\cellcolor[HTML]{EFEFEF}37.75} & \cellcolor[HTML]{EFEFEF}40.69 & \cellcolor[HTML]{EFEFEF}28.89 & \cellcolor[HTML]{EFEFEF}23.63 & \multicolumn{1}{c|}{\cellcolor[HTML]{EFEFEF}36.86}          & \multicolumn{1}{c|}{\cellcolor[HTML]{EFEFEF}24.32} & \multicolumn{1}{c|}{\cellcolor[HTML]{EFEFEF}26.09} & \multicolumn{1}{c|}{\cellcolor[HTML]{EFEFEF}\textbf{43.20}}  & \cellcolor[HTML]{EFEFEF}34.86                \\ \hline
\end{tabular}
}
\caption{Comparison of 10 models on both realistic and semantic top-view maps. Performance is analysed according to four tasks with EM and PM. The best performance for each task is illustrated in \textbf{bold}. }
\label{tab:main results}
\end{table*}

\begin{figure*}[t!]
\centering
\subcaptionbox[Short Subcaption]{%
    Performance with realistic top-view maps 
    \label{fig:rgb fg results}%
}
[%
    0.58\textwidth % width of caption
]%
{%
    \includegraphics[width=0.58\textwidth]{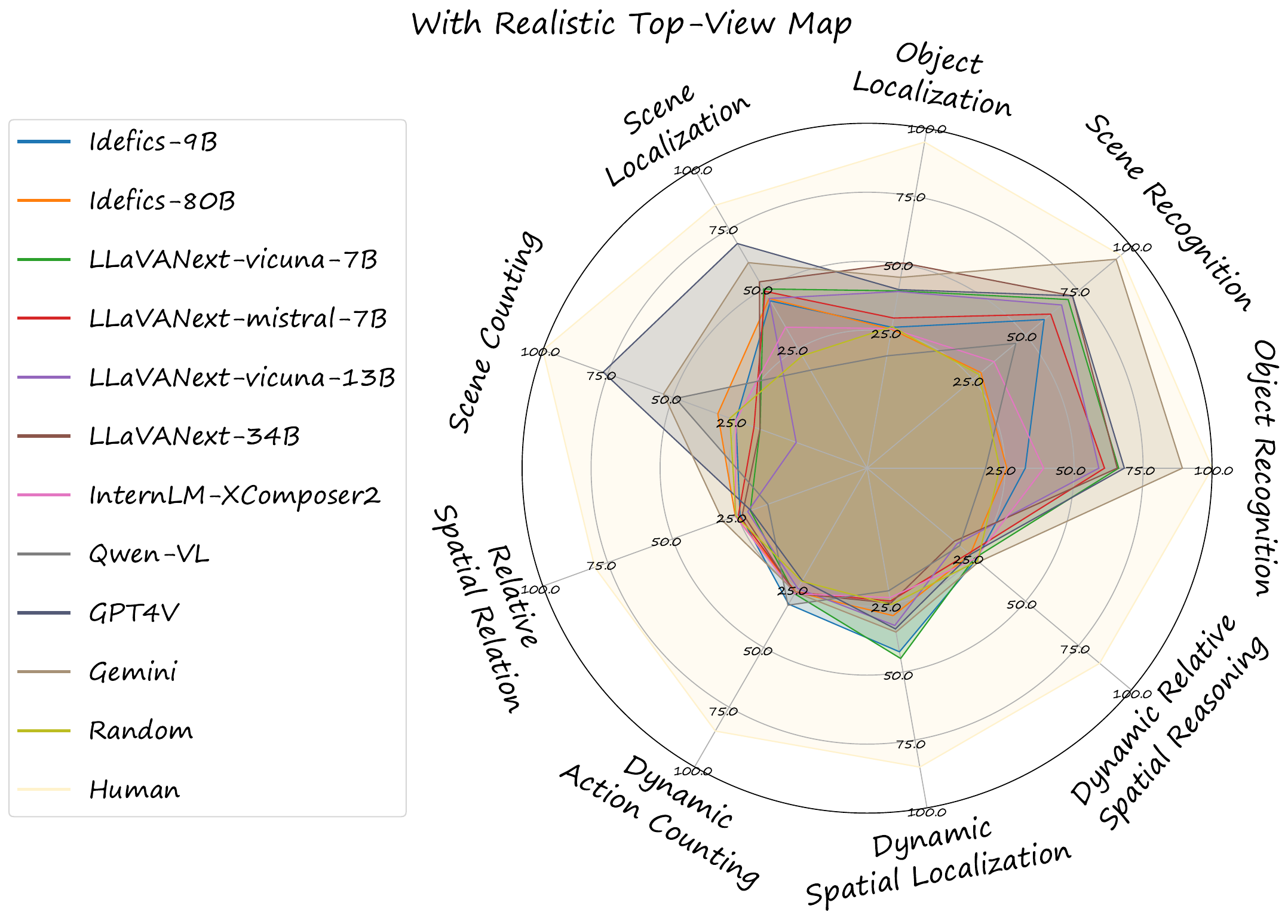}%
}\hfill
\subcaptionbox[Short Subcaption]{%
    Performance with semantic top-view maps 
    \label{fig:semantic fg results}%
}
[%
    0.41\textwidth % width of caption
]%
{%
    \includegraphics[width=0.41\textwidth]{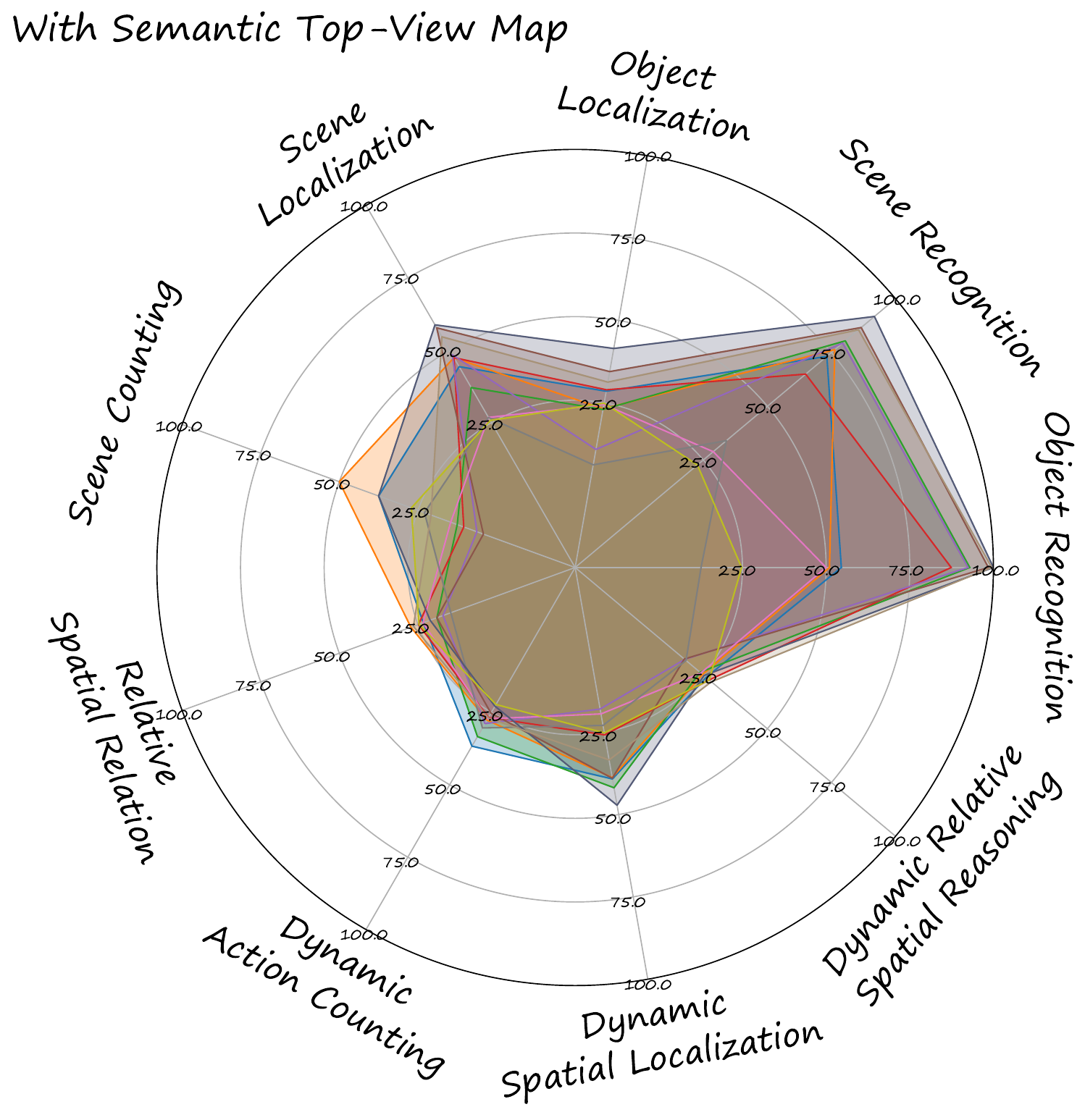}%
}%
\caption[Short Caption]{Visualization of fine-grained comparison with 10 models and human on 9 sub-tasks using realistic and semantic top-view maps, demonstrating that \textit{most current models perform on par with random baseline in spatial reasoning and has a large gap with human performance}. Exact numbers are reported in Table \ref{apptab:fine-grained_results} in the Appendix. }
\vspace{-2mm}
\label{fig:fg results}
\end{figure*}

We first discuss the models' performance across our four tasks, with results summarized in Table \ref{tab:main results}, and fine-grained sub-task performance illustrated in Figure~\ref{fig:fg results}. We find that the performance of current state-of-the-art VLMs is \textit{unsatisfactory} on the proposed \topviewers benchmark with model-wise average EM and PM over all tasks below 50\%. 
Gemini is the best-performing model for realistic maps, while GPT-4V excels in semantic maps. 
For some models, such as Qwen-VL, the results are sometimes much worse than the random baseline. This issue primarily arises from the models' difficulty in following the instructions to choose from the four provided options. 

\rparagraph{Models perform better on recognition and localization tasks compared to reasoning tasks}
Top-View Recognition consistently demonstrates the highest performance across all models. 
Gemini shows human-comparable performance with over 90\% EM score. 
Top-View Localization exhibits lower performance compared to Top-View Recognition, followed by Static Spatial Reasoning. 
The performance difference of various tasks with different levels of complexity underscores \textit{the advantage of our benchmark to capture well-defined and disentangled phenomena}, which allows for controlled studies in controlled environments.

Regarding Dynamic Spatial Reasoning, models perform better on this task than on the previous tasks. Fine-grained performance in Figure \ref{fig:fg results} indicates that the improved performance primarily stems from high accuracy in dynamic action counting and spatial localization, which constitute 18\% and 66\% of the data respectively for this task. We attribute the high accuracy in these areas to the equivalence between navigation path symbols and visual prompting \cite{Shtedritski_2023}. Despite these advancements, the overall EM accuracy remains below 40\%, and \textit{models still struggle with reasoning over dynamic relative spatial relations}.

\rparagraph{Larger models do not always show better spatial awareness}
Surprisingly, our experiment results reveal that larger model sizes do not consistently translate to better performance. 
In Top-View Recognition, closed-source models outperform open-source models by 31.10\% EM with realistic maps and 29.33\% EM with semantic maps. 
However, the performance gap narrows as the task complexity increases.
Using realistic maps as the visual input, Gemini stands out by achieving a minimum of 5.53\% higher EM accuracy in Static Spatial Reasoning compared to other models, while GPT-4V performs worse than Idefics-9B on both Static and Dynamic Spatial Reasoning tasks.
This indicates a lack of significant difference in spatial awareness between closed-source and open-source models for tasks with higher complexity, despite the disparity in their model sizes. 
This trend holds true within open-sourced models as well. 
Both Idefics and LLaVANext model families in some cases show comparable or worse performance with larger sized models than smaller models.  
Similar observations have been made by previous studies \cite{zhong-etal-2021-larger, shi2024need}. 
We conjecture that this might be caused by inadequate evidence of the scaling law \cite{kaplan2020scaling} in the computer vision community \cite{tian2024visual}.
The results on \topviewers thus advocate for further investigation and analysis in this area.

\paragraph{Models perform better in easier tasks with semantic maps.}
In simple tasks such as Top-View Recognition, models generally perform better with semantic maps than with realistic maps, except for Qwen-VL, showing an improvement of 20.35\%. 
However, this advantage decreases in more complex tasks.
For Top-View Localization and Static Spatial Reasoning, models struggle to utilize semantic top-view maps, yielding performances akin to random baselines in both EM and PM accuracy.
One possible explanation is that the semantic top-view image and the input prompt with color-object mapping deviate too much from the models' training data distribution. 
This is further evidenced by the predictions from open-sourced models such as Qwen-VL, which fail to respond to instructions and answer with numbers or RGB values 91.25\% of the time for Top-View Localization and 47.65\% of the time for Static Spatial Reasoning.

\paragraph{Fine-Grained Insights with Sub-Tasks}
Models using realistic maps excel more in the sub-task of Scene Recognition, which involves larger entities, compared to Object Recognition. 
This gap is also evident in a 12.66\% and 19.73\% performance difference between object-level and scene-level localization with both map types.
Conversely, with semantic maps, the model struggles more with scene-level recognition than with realistic maps, showing an 11.09\% lower performance than object-level recognition among closed-source models. 
Most models perform similarly to a random baseline in reasoning over spatial relations but show higher accuracy in scene counting. 
This likely occurs because 95\% of the correct room counts are within a narrow range (1 or 2), reflecting real-life distributions. 
Thus, models leverage commonsense knowledge as the shortcut for counting, as seen in the 54.73\% performance gap (with GPT-4V) between counting scenes and actions. 
However, the spatial localization and reasoning abilities of both open-source and closed-source models still remain unsatisfactory, even at the level of sub-tasks.

\subsection{Further Discussion}

\sparagraph{Gap to Human Performance}
We now study how humans perform on this dataset and the gap between current models and human performance. To this end, we recruited 4 human participants who were not involved in dataset creation for human evaluation.  A total of 60 data points with realistic top-view maps are randomly selected from the sub-tasks, covering all fine-grained question types.\footnote{We did not run human evaluation on semantic maps because they are inherently easier to reason over; they skip the process of recognizing the objects before reasoning, which makes the task simpler but with more sufficient and accurate information for reasoning.} 
We use Fleiss Kappa as the measure of inter-annotator agreement. 
The kappa score is 0.747, indicating substantial agreement shared by the human participants according to \citet{kappa}. 
The average performance of the human participants is shown in Table~\ref{tab:human-eval}, the scores 90\% accuracy across all the sub-tasks.
The experimental results show that there is still a large gap with human performance by over 50\% across all the sub-tasks that involve spatial awareness. 
We also observe that with GPT-4V, human performs 47.78\% higher than the model on average. 
The gap between human and model performance is larger on complex reasoning tasks compared to the recognition tasks, indicating plenty of room for improvement. 

\begin{table}[t]
\centering
\renewcommand\arraystretch{1.05}
\resizebox{\columnwidth}{!}{%
\begin{tabular}{llccc}
\toprule
\textbf{Task}    & \multicolumn{1}{c}{\textbf{Ability}} & \textbf{Size} & \textbf{Human} & \textbf{GPT-4V} \\ \midrule
\multirow{2}{*}{\textbf{TVR}} & Object Recognition                   & 5                  & 95             & 100            \\
                     & Scene Recognition                    & 5                  & 100            & 80             \\ \midrule
\multirow{2}{*}{\textbf{TVL}}  & Object Localization         & 5                  & 95             & 20             \\
                     & Scene Localization          & 10                 & 85             & 60             \\ \midrule
\multirow{2}{*}{\textbf{SSR}}  & Scene Counting                       & 5                  & 100            & 80             \\
                     & Relative Spatial Relation            & 10                 & 80             & 0              \\ \midrule
\multirow{3}{*}{\textbf{DSR}} & Dynamic Action Counting              & 5                  & 85             & 0              \\
                     & Dynamic Spatial Localization         & 10                 & 85             & 40             \\
                     & Dynamic Relative Spatial Reasoning   & 5                  & 85             & 0              \\ \bottomrule
\end{tabular}%
}
\caption{Comparison with human and GPT-4V on all the sub-tasks, demonstrating a huge \textit{gap} between GPT-4V and human. }
\label{tab:human-eval}
\vspace{-6mm}
\end{table}

\rparagraph{Chain-of-Thought Helps Elicit Spatial Reasoning}
Due to the compositionality of Static Spatial Reasoning based on Top-View Recognition and Localization in task design, the model is supposed to answer the question based on the locations of the entities in the top-view map. 
Inspired by this requirement, we explored whether Chain-of-Thought (CoT) reasoning \cite{cot} could facilitate spatial reasoning by initially prompting the model to localize entities before producing the final answer to the question. 
To implement this, we modified the instruction to include:  \textit{``You should first localize the entity and then answer the question based on the locations''}, thereby encouraging the model to process information and think step by step.
Considering that CoT has shown effectiveness in larger models \cite{cot, li-etal-2023-symbolic}, we conducted experiments with GPT-4V and Gemini to evaluate this hypothesis. 
As shown in Table \ref{tab:cot results}, incorporating CoT into the reasoning process notably enhances performance. 
Specifically, the models' accuracy improved by 6.50\% when using realistic maps and 5.14\% with semantic maps. 
This improvement underscores the potential of step-by-step reasoning in enhancing the efficacy of spatial reasoning tasks.

\begin{table}[]
\centering
\renewcommand\arraystretch{1.5}
\resizebox{\columnwidth}{!}{%
\begin{tabular}{l|ccc|ccc}
\hline
\multicolumn{1}{c|}{\textbf{Model}} & \multicolumn{3}{c|}{\textbf{GPT-4V}}                                         & \multicolumn{3}{c}{\textbf{Gemini}}                                   \\
                                    & \textbf{w/o. CoT} & \textbf{w. CoT} & \textbf{$\Delta$}                        & \textbf{w/o.CoT} & \textbf{w. CoT} & \textbf{$\Delta$}                       \\ \hline
\textbf{RGB}                        & 22.16             & 26.74           & {\color[HTML]{34A853} \textbf{+4.58}}  & 31.61            & 40.02           & {\color[HTML]{34A853} \textbf{+8.41}} \\
\rowcolor[HTML]{EFEFEF} 
\textit{Scene Counting}             & \textit{76.74}    & \textit{25.58}  & {\color[HTML]{CB0000} \textit{-51.16}} & \textit{53.49}   & \textit{48.84}  & {\color[HTML]{CB0000} \textit{-4.65}} \\
\rowcolor[HTML]{EFEFEF} 
\textit{Relative Spatial Relations} & \textit{19.82}    & \textit{26.79}  & {\color[HTML]{34A853} \textit{+6.97}}  & \textit{30.68}   & \textit{39.64}  & {\color[HTML]{34A853} \textit{+8.96}} \\ \hline
\textbf{Semantic}                   & 21.73             & 28.07           & {\color[HTML]{34A853} \textbf{+6.34}}  & 26.22            & 30.16           & {\color[HTML]{34A853} \textbf{+3.94}} \\
\rowcolor[HTML]{EFEFEF} 
\textit{Scene Counting}             & \textit{37.50}    & \textit{47.92}  & {\color[HTML]{34A853} \textit{+10.42}} & \textit{20.83}   & \textit{29.17}  & {\color[HTML]{34A853} \textit{+8.34}} \\
\rowcolor[HTML]{EFEFEF} 
\textit{Relative Spatial Relations} & \textit{21.12}    & \textit{27.31}  & {\color[HTML]{34A853} \textit{+6.19}}  & \textit{26.43}   & \textit{30.20}  & {\color[HTML]{34A853} \textit{+3.77}} \\ \hline
\end{tabular}%
}
\caption{Comparison of model performance with or without Chain of Thought (CoT) on Static Spatial Reasoning, showing that \textit{CoT helps elicit spatial reasoning}. }
\label{tab:cot results}
\vspace{-4mm}
\end{table}
\section{Conclusion}
In this study, we designed four tasks to examine the capabilities of VLMs as top-view spatial reasoners, progressing from basic top-view map comprehension to dynamic spatial reasoning along navigation paths.
To facilitate this examination, we collect a natural, high-quality dataset, \topviewers, which includes 11,384 multiple-choice questions, featuring either realistic or semantic top-view maps as the visual input. 
Our extensive experiments involved evaluating 10 VLMs across various model families and sizes on \topviewers. 
The results highlight a critical observation: particularly in complex reasoning tasks, VLMs frequently perform only as well as a random baseline, with even more pronounced deficits when handling tasks with semantic maps. 
Moreover, there is a noticeable performance gap compared to human annotators, underscoring the significant potential for further improvements in this field.
In response to these findings, we discovered that employing chain-of-thought reasoning enhances model performance in spatial reasoning by 5.82\%. 
Despite this progress, the overall performance of VLMs on spatial reasoning remains less than satisfactory.
We hope that our study can set the stage for future research in multimodal spatial reasoning and encourage further investigations into refining the reasoning techniques, moving VLMs closer to human-level proficiency in understanding and reasoning over real-world environments.

\section*{Limitations}

The \topviewers dataset primarily evaluates model performance in entity recognition, localization, and spatial reasoning over 2D top-view maps. 
% task-wise
However, it does not yet include task-oriented planning with spatial awareness, which involves more complex sequential decision-making and dynamic interactions.
Our dataset assumes one correct answer per question, but exploring scenarios with multiple correct answers or no correct answers could further challenge systems and provide valuable insights.
We also advocate for further research to explore how spatial awareness in models impacts downstream tasks such as navigation instruction generation \cite{li2024semantic} and task completion by language agents in real-world environments \cite{pmlr-v229-parashar23a}. 
Moreover, our study is currently limited to 2D top-view maps, whereas spatial reasoning can encompass a variety of modalities and perspectives, such as 3D point clouds.
% model-wise
From the perspective of the models, the rapid progress in VLMs makes it hard to include all new releases such as Idefics 2 \cite{laurenon2024matters}. 
Additionally, multimodal in-context learning (MICL) remains underexplored and is only supported by VLMs trained with interleaved image-text data \cite{baldassini2024makes}. Although not universal across all VLMs, MICL has been effective in handling out-of-distribution tasks \cite{zhang2024out}, which could also be interesting in \topviewers, especially with semantic maps as visual inputs.
In future work, we aim to extend our analysis to include more modalities, evaluate a broader range of models and their capabilities, and investigate additional downstream tasks involving spatial awareness.

\section*{Acknowledgments}
The work was partly supported by a Royal Society University Research Fellowship \textit{Inclusive and Sustainable Language Technology for a Truly Multilingual World'} (no 221137) awarded to Ivan Vuli\'{c}.
We thank Fangyu Liu and Ye Mao for constructive feedback on the draft of this paper. 

\bibliography{anthology,custom}

\appendix

\clearpage
\section{Additional Related Work}
\label{app:related_work}
In addition to Section \ref{sec:related work}, we provide additional related work for the comprehensive understanding of spatial reasoning. 

\rparagraph{Spatial Reasoning on Text}
Spatial reasoning has been investigated with the advancement of LLMs \cite{yamada2024evaluating}. 
Various benchmarks have been proposed to evaluate models' spatial reasoning abilities, including relative spatial relation recognition \cite{DBLP:journals/corr/WestonBCM15, mirzaee-etal-2021-spartqa, stepGame2022shi}, natural language navigation \cite{yamada2024evaluating}, and planning \cite{NEURIPS2023_dc9d5dcf}.
\citet{mirzaee-kordjamshidi-2022-transfer} suggest that introducing synthetic data of spatial reasoning when pre-training helps to improve the spatial awareness of the model. 
\citet{yang-etal-2023-coupling} justify the feasibility of using a logical form as an intermediate representation to improve the spatial reasoning ability in easy scenarios. 
Instead of describing the spatial relations with natural language, \citet{wu2024visualizationofthought} feed the model with a 2D square grid similar to ASCII-art format and prove that visualising the reasoning procedure explicitly helps to improve the model's ability in multi-hop spatial reasoning. 
Constrained by language descriptions, most datasets focus on reasoning over symbols within simple scenarios (\textit{e.g. grid-based navigation}) and are synthetically generated. However, real-life scenarios are often more complex and rich in physical semantics. This raises concerns about the models' actual spatial reasoning abilities compared to their proficiency in understanding linguistic patterns. 
\section{Dataset Construction}
\label{app:dataset construction}

The \topviewers is derived from Matterport3D \cite{Matterport3D} for non-commercial academic use only, under the Term of Use (\href{https://kaldir.vc.in.tum.de/matterport/MP_TOS.pdf}{Matterport End User Licence Agreement For Academic Use of Model Data}). 

In addition to the introduction in Section \ref{sec:dataset}, we provide further details with regard to \topviewers dataset construction. 

\subsection{Top-View Map Construction}
To ensure high-quality top-view map representations, we exclude the 3D environments with low coverage of mesh grids. 
We also prefer environments that are single-floor, in order to avoid the obstruction of objects from different floors. 
After manually going through 90 building-scale 3D environments from Matterport3D \cite{Matterport3D}, we select a total of 7 scenes: \texttt{17DRP5sb8fy, 2azQ1b91cZZ, 2t7WUuJeko7, 5LpN3gDmAk7, EU6Fwq7SyZv, 8WUmhLawc2A, i5noydFURQK}.

\paragraph{Realistic Top-View Map}
We extract realistic top-view maps using MeshLab by placing an orthographic camera on the top of the 3D scenes and taking a camera shot.

\paragraph{Semantic Top-View Map}
We construct the semantic top-view map with Habitat simulation environment \cite{habitat19iccv, szot2021habitat}. 
For each building floor, Matterport3D contains the 2D and 3D semantic segmentation human annotations, which can be retrieved to identify the type of objects as well as the rooms. 
The 3D coordinates of the entity's (object and room) center $(x_i, y_i, h_i)$ and the size of the entity's bounding box $(w_x, w_y, w_h)$ can also be retrieved as part of the circumstantial information. 
This information is then used for the construction of the semantic top-view map. 

When we obtain the object information for the purpose of constructing a top-view semantic map, we design certain rules to exclude specific types of objects from all 40 object annotation categories of Matterport3D. 
We believe these objects could either 1) be less meaningful in terms of semantics or 2) take up a large area in the semantic map, which obstructs other objects beneath. 
The filtered objects include:\texttt{`misc', `ceiling', `objects', `floor', `wall', `void', `curtain', `column', `beam', `board panel'}.

We also filter out the objects based on their heights $h_{obj}$ and sizes $w_{obj}$ compared to the rooms' heights $h_{room}$ and sizes $w_{room}$. 
We only keep the objects if they satisfy the following relations:
$$0.9 \times (h_{room} - \frac{1}{2}w_{room}) \leq h_{obj} - \frac{1}{2} w_{obj}$$
$$1.1 \times (h_{room} + \frac{1}{2}w_{room}) \geq h_{obj} + \frac{1}{2} w_{obj}$$

After having all the object annotations, we use the \texttt{get\_topdown\_map} API of the Habitat simulator to get the top-down map of the scene, which describes the navigable area and the overall shape of the environment, but without any object annotations. 
Based on this map, we then draw the bounding boxes with different colors to represent the objects in the environments. 
Considering that the objects on the top may obstruct the bottom objects in the top-view map, to mimic this characteristic, we create the semantic top-view map based on the heights of the objects, where lower objects are drawn first. 
Table \ref{tab:rgb_objects} shows the mapping between the RGB values and object types used for the creation of a semantic top-view map in our work.

\begin{table}[h!]
\centering
\begin{adjustbox}{max width=\textwidth}
\begin{tabular}{|c|c|}
\hline
\textbf{RGB Values} & \textbf{Label} \\ \hline
\textcolor[RGB]{31, 119, 180}{[31, 119, 180]} & \textcolor[RGB]{31, 119, 180}{void} \\ \hline
\textcolor[RGB]{174, 199, 232}{[174, 199, 232]} & \textcolor[RGB]{174, 199, 232}{wall} \\ \hline
\textcolor[RGB]{255, 127, 14}{[255, 127, 14]} & \textcolor[RGB]{255, 127, 14}{floor} \\ \hline
\textcolor[RGB]{255, 187, 120}{[255, 187, 120]} & \textcolor[RGB]{255, 187, 120}{chair} \\ \hline
\textcolor[RGB]{44, 160, 44}{[44, 160, 44]} & \textcolor[RGB]{44, 160, 44}{door} \\ \hline
\textcolor[RGB]{152, 223, 138}{[152, 223, 138]} & \textcolor[RGB]{152, 223, 138}{table} \\ \hline
\textcolor[RGB]{214, 39, 40}{[214, 39, 40]} & \textcolor[RGB]{214, 39, 40}{picture} \\ \hline
\textcolor[RGB]{255, 152, 150}{[255, 152, 150]} & \textcolor[RGB]{255, 152, 150}{cabinet} \\ \hline
\textcolor[RGB]{148, 103, 189}{[148, 103, 189]} & \textcolor[RGB]{148, 103, 189}{cushion} \\ \hline
\textcolor[RGB]{197, 176, 213}{[197, 176, 213]} & \textcolor[RGB]{197, 176, 213}{window} \\ \hline
\textcolor[RGB]{140, 86, 75}{[140, 86, 75]} & \textcolor[RGB]{140, 86, 75}{sofa} \\ \hline
\textcolor[RGB]{196, 156, 148}{[196, 156, 148]} & \textcolor[RGB]{196, 156, 148}{bed} \\ \hline
\textcolor[RGB]{227, 119, 194}{[227, 119, 194]} & \textcolor[RGB]{227, 119, 194}{curtain} \\ \hline
\textcolor[RGB]{247, 182, 210}{[247, 182, 210]} & \textcolor[RGB]{247, 182, 210}{chest\_of\_drawers} \\ \hline
\textcolor[RGB]{51, 105, 30}{[51, 105, 30]} & \textcolor[RGB]{51, 105, 30}{plant} \\ \hline
\textcolor[RGB]{199, 199, 199}{[199, 199, 199]} & \textcolor[RGB]{199, 199, 199}{sink} \\ \hline
\textcolor[RGB]{188, 189, 34}{[188, 189, 34]} & \textcolor[RGB]{188, 189, 34}{stairs} \\ \hline
\textcolor[RGB]{219, 219, 141}{[219, 219, 141]} & \textcolor[RGB]{219, 219, 141}{ceiling} \\ \hline
\textcolor[RGB]{23, 190, 207}{[23, 190, 207]} & \textcolor[RGB]{23, 190, 207}{toilet} \\ \hline
\textcolor[RGB]{158, 218, 229}{[158, 218, 229]} & \textcolor[RGB]{158, 218, 229}{stool} \\ \hline
\textcolor[RGB]{57, 59, 121}{[57, 59, 121]} & \textcolor[RGB]{57, 59, 121}{towel} \\ \hline
\textcolor[RGB]{82, 84, 163}{[82, 84, 163]} & \textcolor[RGB]{82, 84, 163}{mirror} \\ \hline
\textcolor[RGB]{107, 110, 207}{[107, 110, 207]} & \textcolor[RGB]{107, 110, 207}{tv\_monitor} \\ \hline
\textcolor[RGB]{156, 158, 222}{[156, 158, 222]} & \textcolor[RGB]{156, 158, 222}{shower} \\ \hline
\textcolor[RGB]{99, 121, 57}{[99, 121, 57]} & \textcolor[RGB]{99, 121, 57}{column} \\ \hline
\textcolor[RGB]{140, 162, 82}{[140, 162, 82]} & \textcolor[RGB]{140, 162, 82}{bathtub} \\ \hline
\textcolor[RGB]{181, 207, 107}{[181, 207, 107]} & \textcolor[RGB]{181, 207, 107}{counter} \\ \hline
\textcolor[RGB]{206, 219, 156}{[206, 219, 156]} & \textcolor[RGB]{206, 219, 156}{fireplace} \\ \hline
\textcolor[RGB]{140, 109, 49}{[140, 109, 49]} & \textcolor[RGB]{140, 109, 49}{lighting} \\ \hline
\textcolor[RGB]{189, 158, 57}{[189, 158, 57]} & \textcolor[RGB]{189, 158, 57}{beam} \\ \hline
\textcolor[RGB]{231, 186, 82}{[231, 186, 82]} & \textcolor[RGB]{231, 186, 82}{railing} \\ \hline
\textcolor[RGB]{231, 203, 148}{[231, 203, 148]} & \textcolor[RGB]{231, 203, 148}{shelving} \\ \hline
\textcolor[RGB]{132, 60, 57}{[132, 60, 57]} & \textcolor[RGB]{132, 60, 57}{blinds} \\ \hline
\textcolor[RGB]{173, 73, 74}{[173, 73, 74]} & \textcolor[RGB]{173, 73, 74}{gym\_equipment} \\ \hline
\textcolor[RGB]{214, 97, 107}{[214, 97, 107]} & \textcolor[RGB]{214, 97, 107}{seating} \\ \hline
\textcolor[RGB]{231, 150, 156}{[231, 150, 156]} & \textcolor[RGB]{231, 150, 156}{board\_panel} \\ \hline
\textcolor[RGB]{123, 65, 115}{[123, 65, 115]} & \textcolor[RGB]{123, 65, 115}{furniture} \\ \hline
\textcolor[RGB]{165, 81, 148}{[165, 81, 148]} & \textcolor[RGB]{165, 81, 148}{appliances} \\ \hline
\textcolor[RGB]{206, 109, 189}{[206, 109, 189]} & \textcolor[RGB]{206, 109, 189}{clothes} \\ \hline
\textcolor[RGB]{222, 158, 214}{[222, 158, 214]} & \textcolor[RGB]{222, 158, 214}{objects} \\ \hline
\end{tabular}
\end{adjustbox}
\caption{RGB values and corresponding labels.}
\label{tab:rgb_objects}
\end{table}

After having the top-view maps of the whole floor, we crop them into smaller rooms according to the region boundaries obtained from the Habitat simulator.

\subsection{Structured Question Framework Design}
\label{appsec:question templates}
In order to minimize human labour and standardize the collection pipeline, we adopt the template-based question generation method following the practice of \citet{liu-etal-2023-visual} and design 15 different templates in total to construct the sub-tasks for each task. 
Specifically, we consider benchmarking different perspectives of the model's ability within each task in a fine-grained manner when designing the templates. 
The question templates are also multi-scale in terms of objects or rooms with full or partial top-view maps for Top-View Recognition, Top-View Localization and Static Spatial Reasoning. 
For Dynamic Spatial Reasoning, the designed questions evaluate the recognition and reasoning from the scale of single navigation points (Dynamic Action Counting and Spatial Localization) to the whole path (Dynamic Relative Spatial Reasoning). 

Below we provide the designed templates for all 9 sub-tasks that fall into a total of 4 tasks, with concrete examples shown in Figure \ref{fig:front fig}. 
We also introduce the logic for selecting the correct answer and other wrong choices when constructing the multiple-choice questions. 

\subsubsection{Top-View Recognition}

\begin{table}[h]
    \centering
    \begin{tabular}{p{1.8cm}|p{5cm}}
    \toprule
    \multicolumn{2}{l}{\textbf{Object Recognition}} \\
    \midrule
    Template 1 & Which of the following objects are in the room? \\
    \midrule
    Template 2 & Which of the following objects are not in the room? \\
    \midrule
    \multicolumn{2}{l}{\textbf{Scene Recognition}} \\
    \midrule
    Template 1 & What room is this? \\
    \midrule
    Template 2 & What types of rooms are included in the top-view map below? \\
    \bottomrule
    \end{tabular}
    \caption{Templates for Object and Scene Recognition}
    \label{tab:recognition_tasks}
\end{table}

Table \ref{tab:recognition_tasks} shows the templates we use for the Top-View Recognition task. Considering that some objects and rooms may be hard to recognize from the top view, in addition to the set of filtered objects, we also remove some objects (\texttt{`picture', `mirror', `window', `blinds', `towel', `furniture', `door', `tv\_monitor', `cabinet'}) and rooms (\texttt{`hallway', `entryway/foyer/lobby', `tv'}) when we use the templates to generate questions. 

\subsubsection{Top-View Localization}

\begin{table}[h]
    \centering
    \begin{tabular}{p{1.8cm}|p{5cm}}
    \toprule
    \multicolumn{2}{l}{\textbf{Object Localization}} \\
    \midrule
    Template 1 & Where is the <object> in the top-down map? \\
    \midrule
    \multicolumn{2}{l}{\textbf{Scene Localization}} \\
    \midrule
    Template 1 & Where is the <room> in the top-down map? \\
    \midrule
    Template 2 & What objects does <room> have? \\
    \bottomrule
    \end{tabular}
    \caption{Templates for Object and Scene Localization Tasks}
    \label{tab:spatial_understanding_tasks}
\end{table}

Table \ref{tab:spatial_understanding_tasks} shows the templates for the Top-View Localization task. For the objects, we adopt the range in Top-View Recognition.
For the rooms, we define a set of rooms that are easy and natural to recognize for humans, including: \texttt{`office', `workout/gym/exercise', `kitchen', `bedroom', `dining room', `bar', `balcony', `toilet', `bathroom', `living room', `stairs'}.

\subsubsection{Static Spatial Reasoning}

\begin{table}[h]
    \centering
    \begin{tabular}{p{2cm}|p{5cm}}
    \toprule
    \multicolumn{2}{l}{\textbf{Scene Counting}} \\
    \midrule
    Template 1 & How many <room> are there in the map? \\
    \midrule
    \multicolumn{2}{l}{\textbf{Relative Spatial Relation}} \\
    \midrule
    Template 1 & What's <object1>'s relative spatial relation to <object2>? \\
    \midrule
    Template 2 & What's <room1>'s relative spatial relation to <room2>? \\
    \bottomrule
    \end{tabular}
    \caption{Templates for Scene Counting and Relative Spatial Relation Tasks}
    \label{tab:counting_spatial_relation_tasks}
\end{table}

Table \ref{tab:counting_spatial_relation_tasks} lists the templates for the Static Spatial Reasoning task. For rooms, we restrict the regions within the same range as in Top-View Localization. For the objects, we focus on the objects that are common and big enough to recognize in daily life, which includes: \texttt{`chair', `table', `cushion', `sofa', `bed', `chest\_of\_drawers', `sink', `toilet', `bathtub', `stool', `plant', `stairs', `shower', `fireplace', `gym\_equipment', `seating'}.

\subsubsection{Dynamic Spatial Reasoning}

\begin{table}[h]
    \centering
    \begin{tabular}{p{2cm}|p{5cm}}
    \toprule
    \multicolumn{2}{l}{\textbf{Dynamic Action Counting}} \\
    \midrule
    Template 1 & How many turning <action> are there along the path? \\
    \midrule
    \multicolumn{2}{l}{\textbf{Dynamic Relative Spatial Reasoning}} \\
    \midrule
    Template 1 & Which direction does the navigation path head for? \\
    \midrule
    \multicolumn{2}{l}{\textbf{Dynamic Spatial Localization}} \\
    \midrule
    Template 1 & What rooms does the navigation path pass by? \\
    \midrule
    Template 2 & At which room does the navigation path <action>? \\
    \midrule
    Template 3 & At which object does the navigation path <action>? \\
    \bottomrule
    \end{tabular}
    \caption{Templates for Dynamic Action Counting, Dynamic Relative Spatial Reasoning, and Dynamic Spatial Localization Tasks}
    \label{tab:dynamic_tasks}
\end{table}

For Dynamic Action Counting, we define that a valid turn should involve more than a 30-degree rotation.
For Dynamic Relative Spatial Reasoning, the direction is also defined by the relative spatial relation between the starting point and ending point, where the spatial description is determined by 30 degree intervals. 

\paragraph{Multiple-Choice Question-Answer Pairs}
For the answer to the questions, because we have all the spatial information and semantic annotation of the objects in the scene, we write a set of rules with code for each type of question in order to automatically obtain the golden answer according to the simulation environments. 
For all the wrong choices in the multiple-choice settings, they are randomly chosen from other possible candidates of the same kind (e.g. objects, rooms, numbers, etc.). 
After having all the options for multiple-choice questions, we randomize the order of the options to make the correct choices evenly distributed in ABCD. 

\subsection{Alignment with Human Judgments}
We realize that semantic annotations of environments may sometimes be inaccurate. Moreover, even though we exclude some unreasonable objects, the top views of objects can sometimes be challenging to recognize, even for humans. To address these issues, we have implemented a second stage in our dataset creation process: alignment and verification based on human judgments.

When validating the automatically collected data, the human participants are supposed to check the correctness of the question-answer pair and choose one of the following four actions according to their own judgments:
1) skip the data if it is too bad and strange, 
2) modify the data pair by replacing the options or the entities in the question in order to make it answerable by humans, 
3) correct the answer if it's wrong, 
4) keep the data if it is answerable by humans and correct. 
In order to ensure the quality of the dataset, we communicated to the human participants that they are supposed to be cautious of "accepting" a data point. 
On a practical level, the participants may either discard this data point or modify the options of this data to make the correct choice more distinguishable by humans. 
This helps to exclude the data points where different human judges may diverge and thus ensure the alignment between the dataset and general human judgments. 

We conduct human alignment on Top-View Recognition, Top-View Localization and Static Spatial Reasoning. 
We didn't do human alignment on Dynamic Spatial Reasoning because we fall short of hands in going over the quality of all the data points. 
Therefore, in our experiments, we also provide the corresponding rules of how we obtain the answer for the model with textual description in the prompt (see Appendix \ref{app:prompt}).

\subsection{Dataset Statistics}
\label{appsec:dataset analysis}

\begin{figure*}[htbp]
  \centering 
  \begin{minipage}[t]{0.65\linewidth}
  \centering
  \label{appfig:obj_rgb_dist}\includegraphics[width=\textwidth]{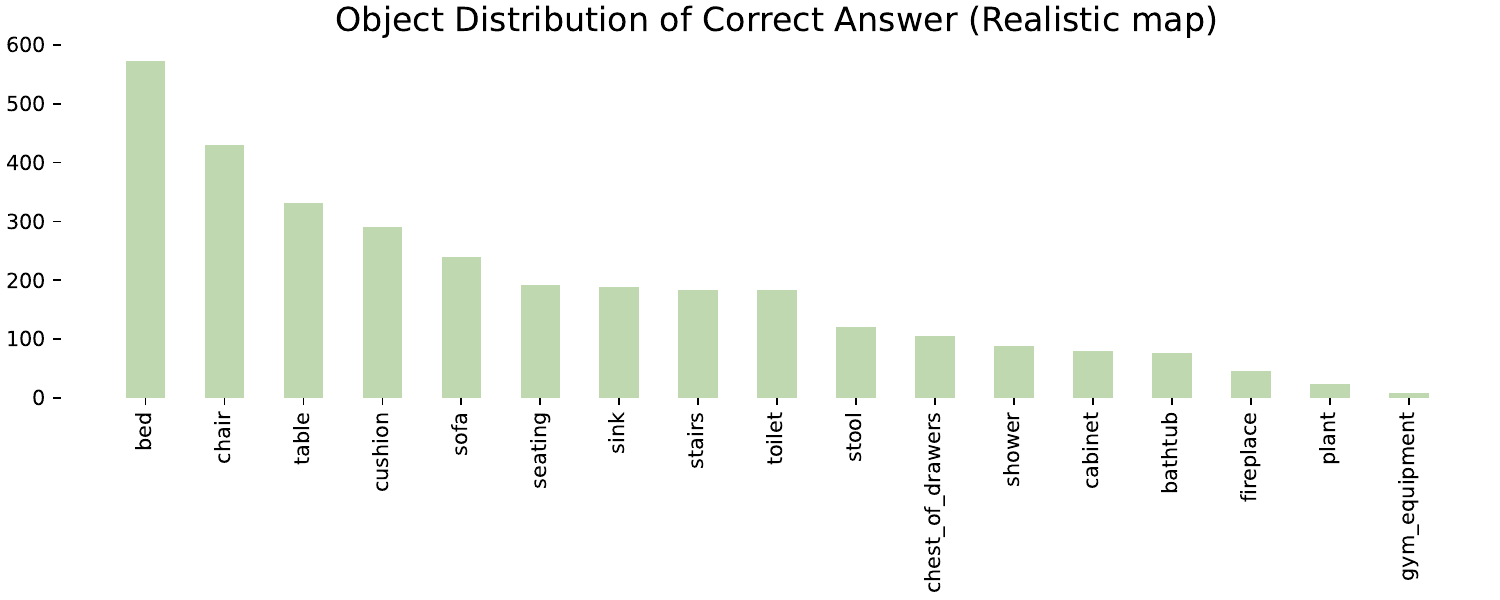}
  \end{minipage}

  \begin{minipage}[t]{0.65\linewidth}
  \centering
  \label{appfig:obj_semantic_dist}\includegraphics[width=\textwidth]{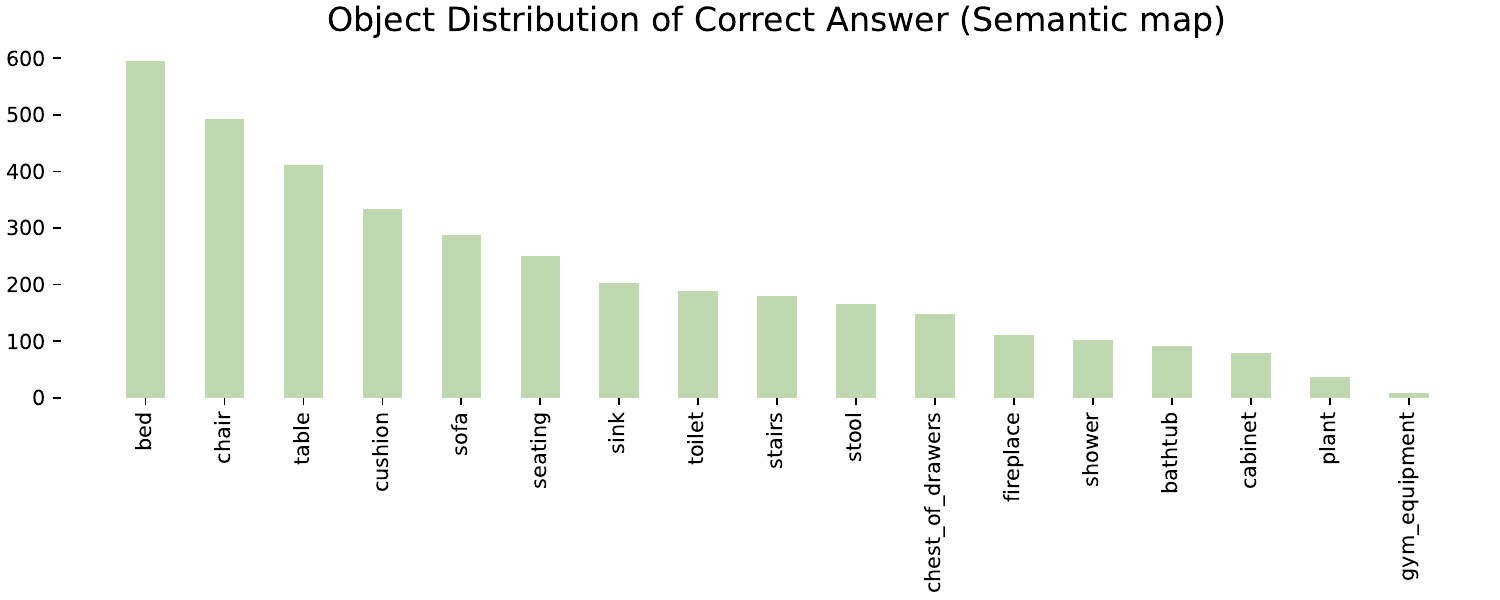}
  \end{minipage}

  \begin{minipage}[t]{0.65\linewidth}
  \centering
  \label{appfig:room_rgb_dist}\includegraphics[width=\textwidth]{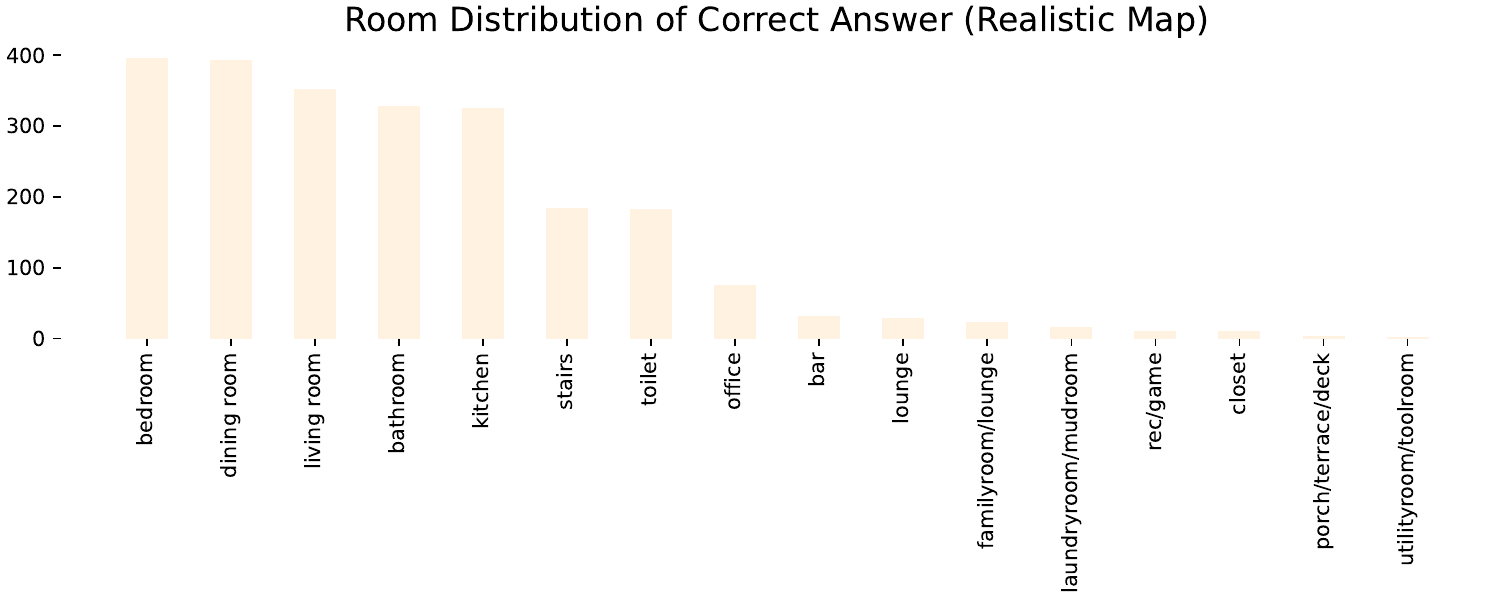}
  \end{minipage}

  \begin{minipage}[t]{0.65\linewidth}
  \centering
  \includegraphics[width=\textwidth]{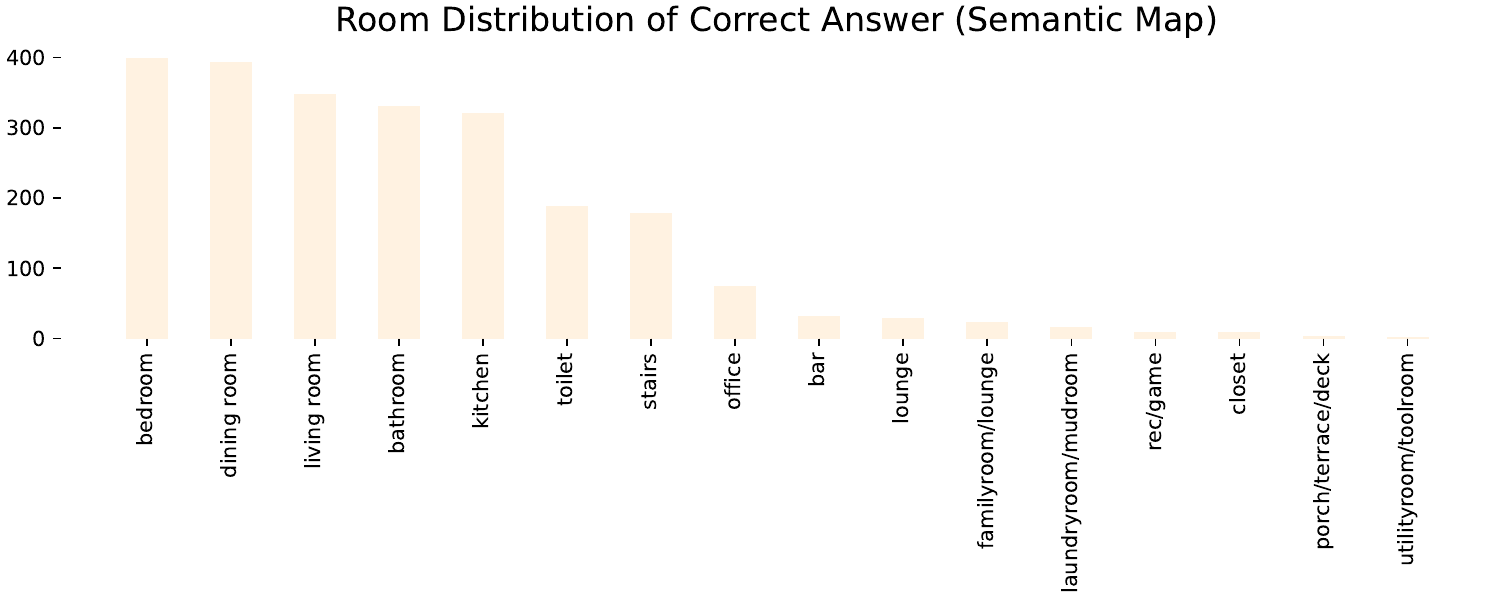}
  \end{minipage}
  
  \caption{Dataset Statistical Analysis}
  \label{appfig:entity-distribution}
\end{figure*}
\begin{table*}
\centering
\renewcommand\arraystretch{1.2}
\resizebox{0.6\linewidth}{!}{%
\begin{tabular}{clcc} 
\hline
\textbf{Task}  & \multicolumn{1}{l}{\textbf{Sub-Task}} & \textbf{Realistic} & \textbf{Semantic}  \\ 
\hline
\textbf{TVR}   & Object Recognition                    & 195          & 198                \\
               & Scene Recognition                     & 97           & 97                 \\ 
\hline
\textbf{TVL}    & Object Localization          & 410          & 470                \\
               & Scene Localization           & 100          & 97                 \\ 
\hline
\textbf{SSR}    & Scene Counting                        & 43           & 48                 \\
               & Relative Spatial Relation             & 1,004        & 1,245              \\ 
\hline
\textbf{DSR}   & Dynamic Action Counting               & 668          & 668                \\
               & Dynamic Spatial Localization          & 2,436        & 2,436              \\
               & Dynamic Relative Spatial Reasoning    & 586          & 586                \\ 
\hline
\textbf{Total} &                                       & 5,539        & 5,845              \\
\hline
\end{tabular}
}
\caption{Distribution of sub-tasks with realistic and semantic top-view maps. }
\label{apptab:sub-task}
\end{table*}

We provide further insights about the datasets with regard to the object and room distribution in Figure \ref{appfig:entity-distribution} and sub-tasks statistics in Table \ref{apptab:sub-task}. 

The visualization demonstrates that the objects or regions that are hard to recognize (\textit{e.g. gym equipment, utility room, etc.}) show fewer occurrence within our dataset compared to those which are easy to identify with little obscure (\textit{e.g. bed, table, bedroom, etc.}). 
\textit{Bed, chair} and \textit{table} are the top-3 most frequently mentioned objects and \textit{bedroom, dining room} and \textit{living room} are the most common regions in the dataset. 
Among all the spatial descriptions, the diagonal spatial relations (\textit{e.g. top right, up left}) are more frequently referred to as the correct choice as relative spatial descriptions in Static Spatial Reasoning while being less frequently used as absolute spatial descriptions in Top-View Localization. 

Regarding the sizes for each sub-task, object-level recognition and localization take a large portion of data in the Top-View Recognition and Localization tasks. 
For Static Spatial Reasoning, reasoning over relative spatial relations takes the main part of the data. 
Dynamic Spatial Localization is the largest task according to the sizes. 
The numbers are different with realistic maps and semantic maps for each task. 
The disparity stems from the second stage of dataset creation, where the human annotators may exclude some data points with realistic maps due to various possible reasons including but not limited to the obscure image or etc.
\section{Experiments}
\label{app:experiment}

\subsection{Inference Parameters}
\begin{table*}
\centering
\arrayrulecolor[rgb]{0.8,0.8,0.8}
\resizebox{0.25\linewidth}{!}{%
\begin{tabular}{cc} 
\arrayrulecolor{black}\hline
\multicolumn{2}{c}{{\cellcolor[rgb]{0.8,0.8,0.8}}Idefics 9B\&80B}           \\ 
\hline
max\_new\_tokens      & 20                                                       \\ 
\hline
\multicolumn{2}{c}{{\cellcolor[rgb]{0.8,0.8,0.8}}LLaVANext 7B\&13B\&34 B}    \\ 
\hline
temperature         & 0                                                        \\ 
\arrayrulecolor[rgb]{0.8,0.8,0.8}\hline
num\_beams          & 1                                                        \\ 
\hline
max\_new\_tokens    & 20                                                       \\ 
\hline
do\_sample          & False                                                    \\ 
\hline
top\_p              & None                                                     \\ 
\arrayrulecolor{black}\hline
\multicolumn{2}{c}{{\cellcolor[rgb]{0.8,0.8,0.8}}XComposer2}                 \\ 
\hline
temperature         & 1                                                        \\ 
\arrayrulecolor[rgb]{0.8,0.8,0.8}\hline
beams               & 5                                                        \\ 
\hline
max\_token          & 20                                                       \\ 
\hline
repetition\_penalty & 1                                                        \\ 
\hline
do\_sample          & False                                                    \\ 
\arrayrulecolor{black}\hline
\multicolumn{2}{c}{{\cellcolor[rgb]{0.8,0.8,0.8}}Qwen-VL}                    \\ 
\hline
max\_new\_tokens      & 20                                                       \\ 
\hline
\multicolumn{2}{c}{{\cellcolor[rgb]{0.8,0.8,0.8}}GPT4V}                      \\ 
\hline
temperature         & 0                                                        \\ 
\arrayrulecolor[rgb]{0.8,0.8,0.8}\hline
max\_tokens         & 1024                                                     \\ 
\hline
img\_size           & 512                                                      \\ 
\hline
img\_detail         & low                                                      \\ 
\arrayrulecolor{black}\hline
\multicolumn{2}{c}{{\cellcolor[rgb]{0.8,0.8,0.8}}Gemini}                     \\ 
\hline
temperature         & 0                                                        \\ 
\arrayrulecolor[rgb]{0.8,0.8,0.8}\hline
max\_tokens         & 1024                                                     \\
\arrayrulecolor{black}\hline
\end{tabular}
}
\caption{Configurations of inference parameters. }
\label{apptab:inference_config}
\end{table*}

We adopt most of the inference parameters for each model from the implementations of VLMEvalKit \cite{2023opencompass}. 
Table \ref{apptab:inference_config} shows the configuration of the inference process for different models. 
If not specified in Table \ref{apptab:inference_config}, we use the default configuration in Huggingface. 

\subsection{Prompt} \label{app:prompt}
Table \ref{apptab:prompt template for main rgb} and \ref{apptab:prompt template for main semantic} show the prompt templates of each task used in the main experiments (Table \ref{tab:main results}) with realistic and semantic top-view maps as visual input respectively. 
Table \ref{apptab:prompt template for cot rgb} and \ref{apptab:prompt template for cot semantic} show the prompt templates used for Chain-of-Thought reasoning using realistic and semantic top-view maps (Table \ref{tab:cot results}). 

Within the prompt templates, \texttt{<QUESTION>} and \texttt{<OPTIONS>} are replaced with the question and option list $O = \{o_0, o_1, o_2, o_3\}$ (\textit{e.g. ``A. bed; B. chair; C. table; D. cushion''}). 
For semantic top-view maps, \texttt{<MAPPING>} is replaced with the RGB-object mapping, as shown below. 
\begin{verbatim}
(196, 156, 148) -> bed
(44, 160, 44) -> door
...
\end{verbatim}

In the task of Dynamic Spatial Reasoning, \texttt{<TASK-SPECIFIC INSTRUCTION>} contains the rules of how we obtain the answer from the simulator for the sub-task Dynamic Action Counting, which is described as follows. 

\begin{verbatim}
Suppose you are a navigation agent tracing 
the path. Your job is to assess whether 
there's a turn at each intermediate point 
and sum up the total turns for the final 
outcome. 
\end{verbatim}

For other sub-tasks in Dynamic Spatial Reasoning, \texttt{<TASK-SPECIFIC INSTRUCTION>} is replaced with an empty string.

\begin{table*}[]
\centering
\begin{tcolorbox}[title = {Realistic Top-View Maps}]
\textit{Top-View Recognition, Top-View Localization and Static Spatial Reasoning}
\\

This is a top-view map of a room. Please respond to the question below by selecting one choice from a list of available options provided. Your response should only include the letter of the chosen option (A, B, C, or D) with no additional explanation.

Question: <QUESTION>

Options: <OPTIONS>;

Answer: 
\tcblower
\textit{Dynamic Spatial Reasoning}
\\

This is a top-view map of a room with the navigation path. The path starts from the green triangle (RGB [0, 255, 0]) and ends at the red star (RGB [255, 0, 0]). The direction of the path is denoted by a series of yellow arrows (RGB [255, 255, 0]), with intermediate points highlighted in RGB [25, 255, 255]. <TASK-SPECIFIC INSTRUCTION> Please respond to the question below by selecting one choice from a list of available options provided. Your response should only include the letter of the chosen option (A, B, C, or D) with no additional explanation.

Question: <QUESTION>

Options: <OPTIONS>;

Answer: 
\end{tcolorbox}
\caption{Prompt templates for main experiments with realistic top-view maps. }
\label{apptab:prompt template for main rgb}
\end{table*}

\begin{table*}[]
\centering
\begin{tcolorbox}[title = {Semantic Top-View Maps}]
\textit{Top-View Recognition, Top-View Localization and Static Spatial Reasoning}
\\

This is a semantic top-view map of a room. Various objects are depicted by colored bounding boxes, each with its corresponding color, and there may be instances of overlap between them. Below are the RGB color codes associated with each object, presented in the format RGB -> Object:

<MAPPING>

Please respond to the question below by selecting one choice from a list of available options provided. Your response should only include the letter of the chosen option (A, B, C, or D) with no additional explanation.

Question: <QUESTION>

Options: <OPTIONS>;

Answer: 
\tcblower
\textit{Dynamic Spatial Reasoning}
\\

This is a semantic top-view map of a room with the navigation path. In the semantic map, various objects are depicted by colored bounding boxes, each with its corresponding color, and there may be instances of overlap between them. The navigation path starts from the green triangle (RGB [0, 255, 0]) and ends at the red star (RGB [255, 0, 0]). The direction of the path is denoted by a series of yellow arrows (RGB [255, 255, 0]), with intermediate points highlighted in RGB [25, 255, 255]. Below are the RGB color codes associated with each object and symbol, presented in the format RGB -> Object:

<MAPPING>

<TASK-SPECIFIC INSTRUCTION> Please respond to the question below by selecting one choice from a list of available options provided. Your response should only include the letter of the chosen option (A, B, C, or D) with no additional explanation.

Question: <QUESTION>

Options: <OPTIONS>;

Answer: 
\end{tcolorbox}
\caption{Prompt templates for main experiments with semantic top-view maps. }
\label{apptab:prompt template for main semantic}
\end{table*}

\begin{table*}[]
\centering
\begin{tcolorbox}[title = {Realistic Top-View Maps}]
\textit{Static Spatial Reasoning}
\\

This is a top-view map of a room. Please respond to the question below by selecting one choice from a list of available options provided. You should explain your reasoning step-by-step by first localizing the entities and then reasoning over the question based on the locations. You should conclude your chosen option (A, B, C, or D) starting with 'The answer is '.

Question: <QUESTION>

Options: <OPTIONS>;

Answer: Let's think step by step. 
\end{tcolorbox}
\caption{Prompt templates for Chain-of-Thought experiments with realistic top-view maps. }
\label{apptab:prompt template for cot rgb}
\end{table*}

\begin{table*}[]
\centering
\begin{tcolorbox}[title = {Semantic Top-View Maps}]
\textit{Static Spatial Reasoning}
\\

This is a semantic top-view map of a room. Various objects are depicted by colored bounding boxes, each with its corresponding color, and there may be instances of overlap between them. Below are the RGB color codes associated with each object, presented in the format RGB -> Object:

<MAPPING>

Please respond to the question below by selecting one choice from a list of available options provided. You should explain your reasoning step-by-step by first localizing the entities and then reasoning over the question based on the locations. You should conclude your chosen option (A, B, C, or D) starting with 'The answer is '.

Question: <QUESTION>

Options: <OPTIONS>;

Answer: Let's think step by step. 
\end{tcolorbox}
\caption{Prompt templates for Chain-of-Thought experiments with semantic top-view maps. }
\label{apptab:prompt template for cot semantic}
\end{table*}

\subsection{Experimental Results}
Table \ref{apptab:fine-grained_results} shows the fine-grained sub-task performance of all the models, which corresponds to Figure \ref{fig:fg results}.

\begin{table*}
\centering
\renewcommand\arraystretch{1.4}
\resizebox{\linewidth}{!}{%
\begin{tabular}{cc|cc|cccc|c|c|c|c} 
\hline
                     & \multicolumn{1}{l|}{}                                   & \multicolumn{2}{c|}{\textbf{Idefics}} & \multicolumn{4}{c|}{\textbf{LLaVANext}}     & \textbf{XComposer2} & \textbf{Qwen-VL} & \textbf{GPT-4V} & \textbf{Gemini}  \\
\multicolumn{2}{l|}{\textbf{Model Size}}                                       & 9B    & 80B                           & vicuna 7B & mistral 7B & vicuna 13B & 34B   & 7B                           & 7B               & API            & API                  \\ 
\hline
\multicolumn{12}{l}{{\cellcolor[rgb]{0.851,0.851,0.851}}\textbf{Realistic Map}}                                                                                                                                                                                      \\ 
\hline
\multirow{2}{*}{TVR} & Object Recognition                                      & 32.31 & 25.64                         & 66.15     & 61.03      & 58.97      & 65.64 & 38.97                        & 17.95            & 68.21          & 89.23                \\
                     & Scene Recognition                                       & 58.76 & 28.87                         & 70.10     & 61.86      & 67.01      & 72.16 & 35.05                        & 45.36            & 72.16          & 92.78                \\ 
\hline
\multirow{2}{*}{TVL}  & Object Localization                                               & 26.83 & 26.10                         & 40.24     & 30.24      & 40.00      & 50.49 & 26.34                        & 16.34            & 40.73          & 45.21                \\
                     & Scene Localization                                                & 45.00 & 46.00                         & 50.00     & 49.00      & 46.00      & 53.00 & 34.00                        & 16.00            & 69.00          & 61.00                \\ 
\hline
\multirow{2}{*}{SSR}  & Scene Counting                                          & 25.58 & 32.56                         & 16.28     & 18.60      & 2.33       & 16.28 & 25.58                        & 48.84            & 76.74          & 53.49                \\
                     & Relative Spatial Relations                              & 24.00 & 25.80                         & 20.02     & 24.60      & 21.02      & 23.01 & 25.80                        & 13.25            & 19.82          & 30.68                \\ 
\hline
\multirow{3}{*}{DSR} & Dynamic Action Counting                                 & 31.89 & 26.80                         & 27.54     & 26.95      & 25.30      & 27.40 & 27.54                        & 32.34            & 22.01          & 26.95                \\
                     & Dynamic Spatial Localization                            & 42.57 & 29.27                         & 45.03     & 23.89      & 32.88      & 24.63 & 22.62                        & 20.11            & 34.15          & 35.30                \\
                     & \multicolumn{1}{l|}{Dynamic Relative Spatial Reasoning} & 26.62 & 23.72                         & 25.77     & 23.04      & 17.58      & 16.21 & 26.11                        & 18.77            & 23.72          & 27.82                \\ 
\hline
\multicolumn{12}{l}{{\cellcolor[rgb]{0.851,0.851,0.851}}\textbf{Semantic Map}}                                                                                                                                                                                 \\ 
\hline
\multirow{2}{*}{TVR} & Object Recognition                                      & 54.55 & 51.01                         & 92.93     & 87.37      & 92.42      & 98.48 & 50.00                        & 12.63            & 100.00         & 99.49                \\
                     & Scene Recognition                                       & 73.20 & 76.29                         & 80.41     & 64.95      & 79.38      & 86.60 & 28.87                        & 34.02            & 91.75          & 85.57                \\ 
\hline
\multirow{2}{*}{TVL}  & Object Localization                                               & 28.51 & 23.19                         & 22.98     & 28.94      & 10.85      & 34.47 & 24.47                        & 6.17             & 41.49          & 31.28                \\
                     & Scene Localization                                                & 44.33 & 47.42                         & 37.11     & 47.42      & 48.45      & 57.73 & 26.80                        & 26.80            & 58.76          & 54.64                \\ 
\hline
\multirow{2}{*}{SSR}  & Scene Counting                                          & 37.50 & 50.00                         & 12.50     & 10.42      & 6.25       & 4.17  & 14.58                        & 22.92            & 37.50          & 20.83                \\
                     & Relative Spatial Relations                              & 23.29 & 27.23                         & 18.96     & 24.82      & 17.03      & 18.96 & 23.37                        & 14.54            & 21.12          & 26.43                \\ 
\hline
\multirow{3}{*}{DSR} & Dynamic Action Counting                                 & 36.68 & 27.69                         & 33.38     & 26.80      & 28.89      & 25.30 & 27.25                        & 30.39            & 22.90          & 26.95                \\
                     & Dynamic Spatial Localization                            & 39.20 & 38.79                         & 41.87     & 25.82      & 17.98      & 39.00 & 19.46                        & 22.95            & 47.17          & 33.42                \\
                     & \multicolumn{1}{l|}{Dynamic Relative Spatial Reasoning} & 26.11 & 24.74                         & 23.72     & 27.30      & 17.75      & 17.58 & 24.06                        & 18.26            & 25.26          & 28.16                \\
\hline
\end{tabular}
}
\caption{Fine-grained results of 10 VLMs on different sub-tasks, corresponding to the visualization in Figure \ref{fig:fg results}. }
\label{apptab:fine-grained_results}
\end{table*}

\end{document}